\documentclass[sigconf]{acmart}
\AtBeginDocument{%
  }

\usepackage{graphicx}
\usepackage{amsmath}
\usepackage{booktabs}
\usepackage{xpatch}
\usepackage{array}
\usepackage{multirow}
\usepackage{xcolor,colortbl}
\usepackage{enumitem}
\usepackage{bbm}
\usepackage{pifont}
\usepackage{booktabs}
\usepackage{cuted}
\usepackage{makecell}
\usepackage{subcaption} 

\definecolor{Blue-Gray}{RGB}{72, 116, 203}

\newcommand{\cmark}{\textcolor{green!60!black}{\ding{51}}}
\newcommand{\xmark}{\textcolor{red}{\ding{55}}}

\copyrightyear{2026}
\acmYear{2026}
\setcopyright{cc}
\setcctype{by-nc-nd}
\acmConference[MM '26]{Proceedings of the 34th ACM International Conference on Multimedia}{November 10--14, 2026}{Rio de Janeiro, Brazil}
\acmBooktitle{Proceedings of the 34th ACM International Conference on Multimedia (MM '26), November 10--14, 2026, Rio de Janeiro, Brazil}
\acmDOI{10.1145/3767308.3835950}
\acmISBN{979-8-4007-2213-4/2026/11}

\begin{document}

\title[HybridWorldSim]{HybridWorldSim: A Scalable and Controllable High-fidelity Simulator for Autonomous Driving}

\newcommand{\HybridWorldSimAuthorBlock}{%
Qiang Li$^{1,*}$,
Yingwenqi Jiang$^{1,2,*}$,
Tuoxi Li$^{1}$,
Duyu Chen$^{1}$,
Xiang Feng$^{1,2}$,
Yucheng Ao$^{3}$,
Shangyue Liu$^{1}$,\\
Xingchen Yu$^{3}$,
Youcheng Cai$^{3}$,
Yumeng Liu$^{3,\dagger}$,
Yuexin Ma$^{2,\dagger}$,
Xin Hu$^{1,\ddagger}$,
Li Liu$^{1}$,
Yu Zhang$^{1}$,\\
Linkun Xu$^{1}$,
Bingtao Gao$^{1}$,
Xueyuan Wang$^{1}$,
Shuchang Zhou$^{1}$,
Xianming Liu$^{1}$,
Ligang Liu$^{3}$\\[0.3em]
{\normalsize
$^{1}$XPeng Inc.
\quad
$^{2}$ShanghaiTech University
\quad
$^{3}$University of Science and Technology of China}\\[0.2em]
{\normalsize
$^{*}$Equal contribution.
\quad
$^{\dagger}$Corresponding author.
\quad
$^{\ddagger}$Project leader.}%
}

\author{\HybridWorldSimAuthorBlock}

\makeatletter
\gdef\authors{%
Qiang Li\and
Yingwenqi Jiang\and
Tuoxi Li\and
Duyu Chen\and
Xiang Feng\and
Yucheng Ao\and
Shangyue Liu\and
Xingchen Yu\and
Youcheng Cai\and
Yumeng Liu\and
Yuexin Ma\and
Xin Hu\and
Li Liu\and
Yu Zhang\and
Linkun Xu\and
Bingtao Gao\and
Xueyuan Wang\and
Shuchang Zhou\and
Xianming Liu\and
Ligang Liu%
}
\gdef\shortauthors{Qiang Li et al.}
\makeatother


\begin{abstract}
Realistic and controllable simulation is critical for advancing end-to-end autonomous driving, yet existing approaches often struggle to support novel view synthesis under large viewpoint changes or to ensure geometric consistency. We introduce \textbf{HybridWorldSim}, a hybrid simulation framework that integrates multi-traversal neural reconstruction for static backgrounds with generative modeling for dynamic agents. This unified design addresses key limitations of previous methods, enabling the creation of diverse and high-fidelity driving scenarios with reliable visual and spatial consistency. To facilitate robust benchmarking, we further release a new multi-traversal dataset \textbf{MIRROR} that captures a wide range of routes and environmental conditions across different cities. Extensive experiments demonstrate that HybridWorldSim surpasses previous state-of-the-art methods, providing a practical and scalable solution for high-fidelity simulation and a valuable resource for research and development in autonomous driving.
\end{abstract}



\begin{teaserfigure}
    \centering
\vspace{-13pt}
  \includegraphics[width=0.95\textwidth]{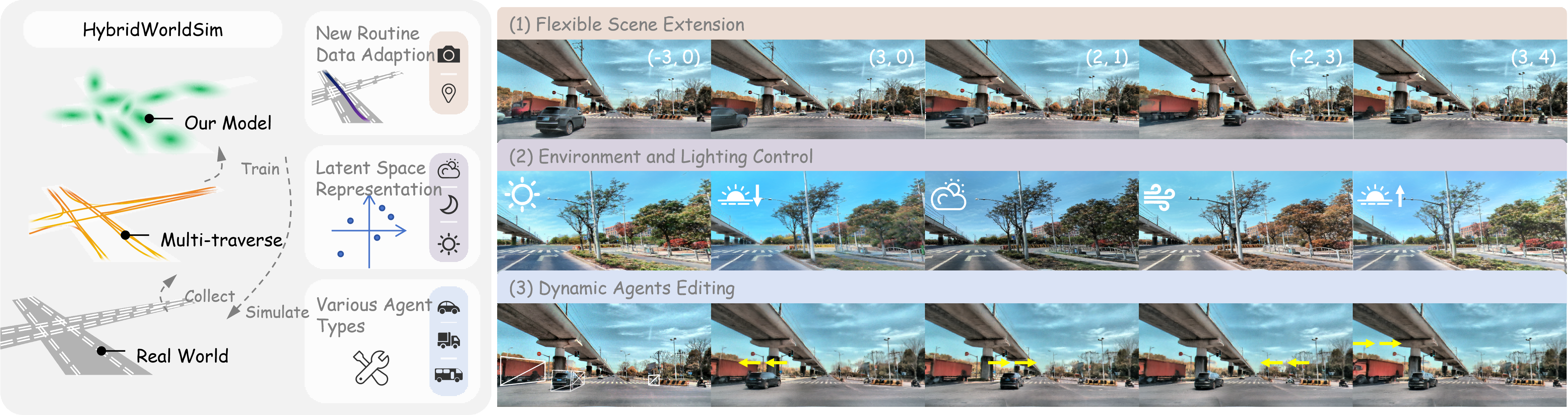}
  \vspace{-10pt}
  \caption{
  We introduce \textbf{HybridWorldSim}, a scalable simulator that couples multi-trajectory neural reconstruction for static backgrounds with generative modeling for dynamic agents.
    }
  \label{fig:teaser}
\end{teaserfigure}




\maketitle

\section{Introduction}
\label{sec:intro}




End-to-end driving models, which directly map sensor inputs to driving actions, have made remarkable progress in recent years and are now regarded as a promising paradigm for autonomous vehicles~\cite{chen2023e2esurvey,chib2023recent}. However, these models critically depend on access to diverse and realistic data that captures the complexity of real-world environments. Collecting and annotating such data in the physical world is both costly and time-consuming, often failing to cover rare or safety-critical scenarios~\cite{sun2020scalability,caesar2020nuscenes}. As a result, high-fidelity closed-loop driving simulators have become indispensable for large-scale training and evaluation~\cite{yang2023unisim,jia2024bench2drive,zhou2024hugsim}, yet achieving both realism and flexibility in these simulators remains an open challenge.

Existing closed-loop simulators generally fall into three technical paradigms, each with significant limitations. First, fully virtual 3D environments, such as CAD-based systems like CARLA~\cite{Dosovitskiy2017CARLA} and game engines like GTA5~\cite{Richter_2016_ECCV} , provide perfect controllability and annotation, but their synthetic nature introduces a notable gap from real-world complexity. Second, neural rendering-based or 3DGS-based reconstruction methods~\cite{yan2024street,chen2023periodic,S-NeRF++,chen2025omnire}, which achieve photorealistic novel-view synthesis and support the separate modeling of static backgrounds and dynamic objects. However, these methods have difficulty adapting to changes in illumination and are not easily extendable to incorporate dynamic agents, particularly vehicles with uncommon shapes or appearances. Third, recent advances in conditional video generation synthesize entire driving scenes from semantic or trajectory inputs~\cite{gao2024vista,wen2024panacea,gao2025magicdrive-v2}, offering controllability and diversity, but often suffering from geometric inconsistency, degraded realism, and poor temporal coherence.

To address these challenges, we propose \textbf{HybridWorldSim}, a novel simulation framework that unifies the strengths of scene reconstruction and generative modeling. Leveraging only vision-based multi-traversal data, our approach enables robust and high-fidelity reconstruction of complex driving scenes using a 3DGS-based method~\cite{kerbl20233dgs}, which efficiently fuses observations from multiple traversals under varying conditions for improved completeness and geometric consistency, as shown in \autoref{fig:teaser}. Building on the reconstructed 3D scene, we further introduce a diffusion-based multi-view video generation pipeline that synthesizes dynamic vehicles on top of geometry-consistent novel-view renderings, rather than generating entire scenes from scratch. By anchoring dynamic generation to the rendered static background and scene geometry, our framework simultaneously preserves visual realism, temporal coherence, and cross-view consistency.

Unlike approaches such as HUGSIM~\cite{zhou2024hugsim}, which rely on pre-collected 3D assets and require explicit modeling or reconstruction for every new dynamic agent, our framework can flexibly incorporate new dynamic observations directly from camera data. This improves the scalability of simulator extension: newly collected data can enrich both the appearance diversity of reconstructed environments and the diversity of dynamic agents and behaviors, without requiring per-agent 3D asset construction. As more multi-traversal data becomes available for an already reconstructed region, the simulator can be continuously expanded with richer traffic content while retaining the same underlying static scene representation. Taken together, these strengths make our paradigm a controllable, scalable, and high-fidelity closed-loop driving simulator.

While our simulation pipeline is compatible with existing auto-driving datasets, we further introduce a dataset, \textbf{MIRROR} (\textbf{M}ulti-traversal \textbf{I}mmersive \textbf{R}oad-scene \textbf{R}ecordings for \textbf{O}pen-world simulato\textbf{R}), which serves as a complementary resource to the simulator. Our dataset provides multi-traversal coverage of diverse traffic scenarios and times of day, collected across 6 cities using 7 types of production vehicles with real human driving patterns. It contains 10 distinct scenes, covering a total area of 1.25 km$^2$. The dataset offers over 2 hours of driving data, captured with 7 cameras under day, night, and rain conditions.
Furthermore, both our data collection process and simulation framework rely solely on RGB cameras, eliminating the need for Lidar or radar sensors. This significantly reduces costs and makes it easier to scale the dataset.

In summary, our main contributions are:
\begin{itemize}
\item We develop \textbf{HybridWorldSim}, a scalable and controllable closed-loop simulator for driving scenarios using only camera data. Our system supports high-fidelity free-viewpoint navigation, scene appearance editing, and flexible control of dynamic agents, enabling photorealistic simulation at urban scale.
\item We release, \textbf{MIRROR}, a multi-traversal dataset, covering diverse scenarios and lighting conditions, collected by various production vehicles using only RGB cameras.
\item We propose a unified approach for vision-based 3D scene reconstruction and generative dynamic modeling, achieving state-of-the-art performance on both public benchmarks and our \textbf{MIRROR} dataset. 
\end{itemize}
Our work paves the way for continuously improving, real-world-scale driving simulators, significantly advancing research and development in end-to-end autonomous driving.

\begin{figure*}[htp]
    \centering
    \includegraphics[width=\textwidth]{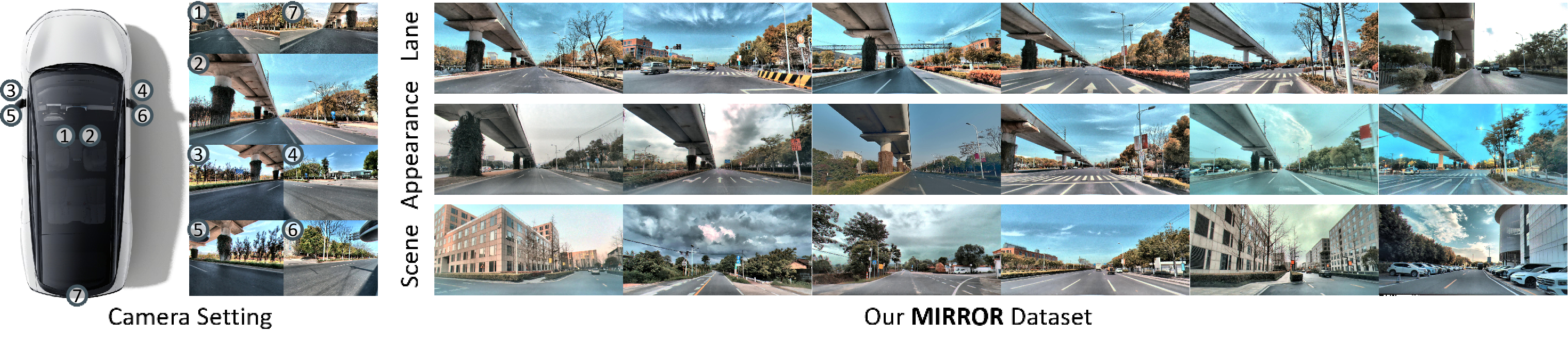}
    \vspace{-3em}
    \captionof{figure}{
    We present our multi-traversal driving dataset \textbf{MIRROR}, collected using various mass-production vehicles, each equipped with a standardized seven-camera rig providing 360-degree coverage. 
    MIRROR dataset captures realistic driving patterns through naturalistic driving behaviors, demonstrates rich multi-traversal diversity with repeated passes through identical regions, and encompasses diverse environmental conditions including varying weather and illumination. 
    }
    \vspace{-1em}
    \label{fig:dataset}
\end{figure*}

\section{Related Work}
\label{sec:related_work}

\subsection{Simulators for Autonomous Driving}
Early progress in autonomous driving simulation focus on synthetic data generation and traffic simulation, such as game-based dataset creation~\cite{Richter_2016_ECCV} and microscopic traffic modeling~\cite{behrisch2011sumo}. Building on these foundations, simulators such as CARLA~\cite{Dosovitskiy2017CARLA} and AirSim~\cite{airsim2017fsr} offered flexible, open environments with configurable weather, traffic, and sensor suites, supporting research in end-to-end policy learning and sensor fusion. More recent works further improve fidelity and realism with complex agent behaviors, and closed-loop interactions~\cite{caesar2021nuplan}.
In addition to simulators built on synthetic environments, recent research increasingly focuses on constructing simulators directly from real-world data for enhanced realism and diversity. For instance, HugSim~\cite{zhou2024hugsim}, OmniRe~\cite{chen2025omnire} leverages 3DGS to reconstruct urban scenes from realistic driving data, enabling simulation grounded in real environments. 
Complementary to reconstruction-based approaches, recent world models explore driving simulation through video generation~\cite{long2025survey}. Pixel-space generative models~\cite{gao2023magicdrive,yang2024drivearena,wen2024panacea,gao2025magicdrive-v2,wang2023drivedreamer,zhao2025drivedreamer,wang2025stage, zheng2026x, zhang2025epona} synthesize diverse driving scenarios directly, while hybrid methods~\cite{gao2024magicdrive3d,yan2025renderworld,zhao2025drivedreamer4d} incorporate 3D representations to improve controllability and geometric consistency. These advances broaden the toolkit for creating realistic and flexible simulation environments.

\subsection{Scene Reconstruction}
Accurate scene reconstruction is fundamental for autonomous driving, supporting perception, localization, and simulation. 
Traditional reconstruction methods use multi-view stereo (MVS) and structure-from-motion (SfM) techniques~\cite{hiep2009towards, yao2018mvsnet,schonberger2016structure,chen2019point} to build driving scenes, allowing for the 3D modeling of urban environments as combinations of meshes and geometric primitives~\cite{lafarge2012hybrid}. However, these methods require laborious data processing, and suffer from lower photorealism, limited view interpolation capability. 
Recent developments in novel-view synthesis address these limitations. Neural volumetric representations, most notably NeRF~\cite{mildenhall2021nerf} and its extensions~\cite{barron2021mip,InstantNGP}, have achieved impressive results in rendering high-quality, realistic views of static scenes. However, these methods are typically limited to static scenes and require intensive training. To address efficiency and scalability, 3DGS~\cite{kerbl20233dgs} has emerged as an explicit and efficient alternative for scene reconstruction and rendering, finding increasing application in large-scale urban and dynamic scenarios\cite{chen2023periodic, yan2024street, Wei_2025_ICCV,peng2025desire}. Nonetheless, 3DGS models often exhibit artifacts when viewpoints are sparse, which is a common situation in autonomous driving data.
To overcome this, multi-traversal data collection and reconstruction strategies such as MTGS~\cite{li2025mtgs} have been proposed to enhance scene completeness and temporal consistency.

\subsection{Driving Scene Editing}
Driving scene editing encompasses both global appearance modification and the insertion or alteration of dynamic foreground vehicles in urban environments.
3D-based methods achieve these goals by editing reconstructed geometry or object assets. Recent advances in relightable 3D Gaussian Splatting~\cite{wu2024deferredgs,kaleta2025lumigauss} enable realistic relighting and material editing, while CAD-based approaches~\cite{engelmann2017samp,wang2023cadsim,uy2020deformation,avetisyan2019scan2cad,gumeli2022roca,lu2024urbancad,zhou2024hugsim,du20243drealcar} and NeRF-based methods~\cite{S-NeRF++} support foreground editing, but often suffer from limited flexibility and unnatural blending.
2D-based methods instead offer greater flexibility and visual quality. GAN-based approaches~\cite{regmi2018cross,lin2020gan,swerdlow2024street,zhang2018deeproad} are effective for style transfer and global appearance editing but struggle with object-specific modifications. Recently, diffusion models~\cite{zhu2025scenecrafter,wei2024editable,liang2024driveeditor} have emerged as powerful tools for both global and localized scene editing, offering greater flexibility and improved visual fidelity, but often struggle with geometric consistency.
Some recent works integrate diffusion model with 3D priors for scene editing~\cite{urbangiraffe,gao2024magicdrive3d,wu2025difix3d}. Motivated by this, we propose a hybrid strategy that unifies 3D structure and 2D editing for realistic and versatile driving scene editing.




\begin{table*}[tbp]
\footnotesize
\begin{center}
\caption{
Comparison of auto-driving datasets. 
\textbf{Multiple Traversals}: Whether the dataset contains multiple traversals in the same scene; 
\textbf{Real Driving Pattern}: Whether the ego-vehicle's driving pattern matches real human driving habits; 
\textbf{Real Scene}: Whether the data is collected in real-world scenes; 
\textbf{Area}: Total covered area; 
\textbf{\#Scenes}: Number of distinct scenes; 
\textbf{Avg. Area per Scene}: Average covered area per scene; 
\textbf{\#City}: Number of cities; 
\textbf{\#Hours}: Number of driving hours; 
\textbf{\#Cam}: Number of cameras; 
\textbf{Light}: Lighting conditions (e.g., Day/Night/Rain). 
``--'' denotes not available.
}
\label{tab:dataset}
\vspace{-1.5em}

\resizebox{\linewidth}{!}{%
\begin{tabular}{lcccccccccc}
\toprule
Dataset & 
\makecell{Multiple\\Traversals} & 
\makecell{Real\\Driving\\Pattern} & 
\makecell{Real\\Scene} & 
\makecell{Area\\({\scriptsize km$^2$})} & 
\makecell{\#Scenes} & 
\makecell{Avg. Area\\per Scene\\({\scriptsize $\times 10^{-3}$ km$^2$)}} & 
\makecell{\#City} & 
\makecell{\#Hours} & 
\makecell{\#Cam} & 
Light \\
\midrule
KITTI~\cite{geiger2013vision}         & \xmark & \xmark & \cmark & --    & 22    & --     & 1   & 1.5   & 4    & Day   \\
NuScenes~\cite{caesar2020nuscenes}    & \xmark & \xmark & \cmark & 5     & 1000  & 5  & 4   & 5.5   & 6    & Day/Night   \\
Argo~\cite{chang2019argoverse}        & \xmark & \xmark & \cmark & 1.6   & 113   & 14  & 2   & 1     & 9    & Day/Night   \\
Waymo~\cite{sun2020scalability}       & \xmark & \xmark & \cmark & 76    & 1150  & 66  & 3   & 6.4   & 5    & Day/Night   \\
nuPlan~\cite{caesar2021nuplan}        & \cmark & \cmark & \cmark & --     & --     & --     & 4   & 1282  & 8    & Day/Rain   \\
Para-Lane~\cite{ni2025lane}           & \cmark & \xmark & \cmark & --     & 25    & --     & 5   & 0.5   & 5    & Day   \\
Open Mars~\cite{li2024multiagent}     & \cmark & \xmark & \cmark & 0.53  & 66    & 8  & 1   & 40    & 6    & Day/Night    \\
XLD~\cite{li2025xld}                  & \cmark & \xmark & \xmark & --     & 6     & --     & --   & --     & 3    & Day/Rain    \\
Ours                                  & \cmark & \cmark & \cmark & 1.25  & 10    & 125  & 6   & 2     & 7    & Day/Night/Rain    \\
\bottomrule
\end{tabular}
}
\end{center}
\vspace{-1.5em}

\end{table*}

\section{Dataset}

The development of controllable and scalable high-fidelity simulators for autonomous driving requires datasets that go beyond scale and diversity, offering detailed coverage of real-world driving patterns and scene dynamics. Although existing benchmarks have greatly advanced the field, key aspects of real-world driving remain insufficiently represented. Therefore, we propose the MIRROR dataset, which addresses these gaps with three key characteristics: \textbf{(1) Realistic Driving Patterns}, \textbf{(2) Multi-Traversal Diversity}, and \textbf{(3) Diverse Environmental Conditions}. A detailed comparison with existing autonomous driving datasets is provided in \autoref{tab:dataset}. The MIRROR dataset establishes a richer and more versatile foundation for high-fidelity simulation. Below, we elaborate on each key feature that we aim to achieve.

\noindent\textbf{Realistic Driving Patterns.}
In contrast to existing datasets, which are typically collected using dedicated vehicles operating at relatively low speeds, our dataset is sourced from real-world user driving sessions. This approach captures authentic vehicle speeds across a broad spectrum of urban and suburban environments. As a result, the data more accurately reflects real driving behaviors and conditions, making it particularly well-suited for training and evaluating deblurring algorithms and other perception tasks under realistic scenarios. 

\noindent\textbf{Multi-Traversal Coverage.}
Most existing driving datasets~\cite{geiger2013vision,caesar2020nuscenes,chang2019argoverse,sun2020scalability} contain only a single traversal for each route or scene. This limitation severely restricts their usefulness for high-fidelity simulation: when the simulated vehicle’s trajectory deviates significantly from the original recorded path, the quality of the rendered environment and dynamic objects drops sharply, leading to unrealistic or incomplete simulation results. In contrast, nuPlan~\cite{caesar2021nuplan} and our MIRROR provide multiple traversals of the same routes, with MIRROR offering even higher traversal counts per scene, and supporting further extension with additional data as needed. This multi-traversal setting supplies abundant observations of each location under varied conditions, enabling more robust and accurate modeling of both static scenes and dynamic objects. As a result, our dataset is particularly valuable for constructing high-quality driving simulators and developing algorithms that require reliable scene and object representations under diverse driving behaviors. 

\noindent\textbf{Diverse Environmental Conditions.}
While some recent datasets, such as nuPlan~\cite{caesar2021nuplan} and Para-Lane~\cite{ni2025lane}, also provide multi-traversal data along the same routes, the intervals between traversals are typically short and data are collected during daytime under relatively consistent conditions, resulting in limited appearance variation. The Open MARS Dataset~\cite{li2024multiagent} also offers multi-traversal data with varying lighting, with each traversal covering a reasonable duration. However, the spatial coverage per location is limited, as data collection is restricted to 67 locations along the driving route, each spanning a circular area with a 50-meter radius. In contrast, our dataset defines much larger regions of interest with a 200-meter radius for each location, enabling significantly broader spatial coverage and more comprehensive viewpoint diversity for scene reconstruction.
Besides, our dataset also encompasses a wide range of weather, lighting, and times of day. This results in significantly broader appearance diversity, better reflecting real-world complexity and facilitating robust appearance modeling, comprehensive evaluation, and research on generalization and domain adaptation.

To enable these features, we design a comprehensive data collection pipeline, deploying 7 types of production vehicles equipped with advanced onboard sensors, as illustrated in \autoref{fig:dataset}. We strategically select GPS points across the map as centers of 200-meter-radius regions of interest (ROIs), aiming to capture a wide range of urban and suburban environments. Data recording is automatically triggered whenever a vehicle’s trajectory passes through an ROI, provided the trajectory lasts at least 10 seconds, covers no less than 20 meters, and maintains over 90\% overlap with the ROI. This fully automated and systematic process ensures rich, diverse, and realistic coverage of driving behaviors and scene dynamics, faithfully reflecting the complexity of real-world traffic scenarios.


\section{Method}
\begin{figure*}[h]
    \centering
    \includegraphics[width=1.0\textwidth]{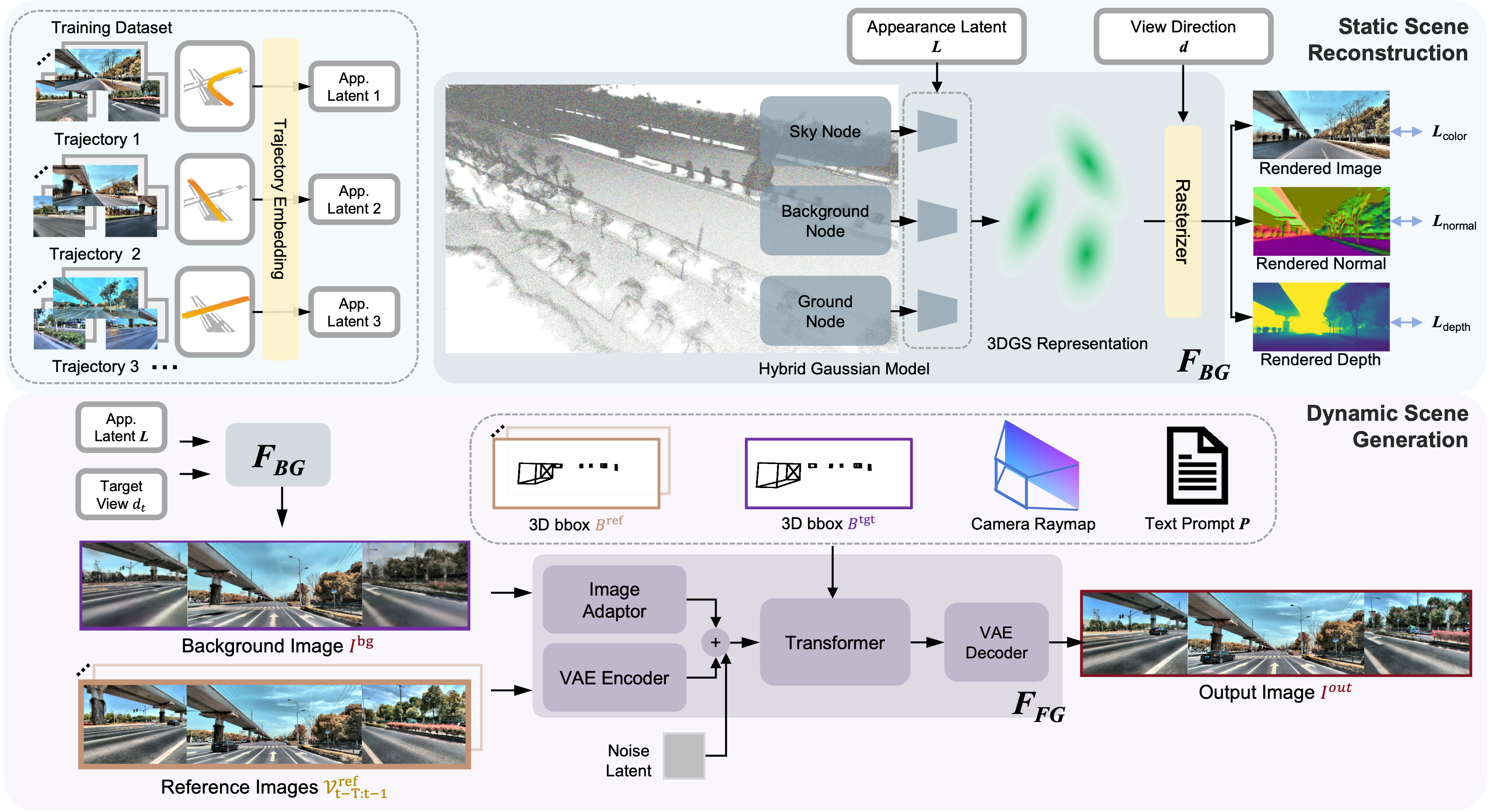}
    \vspace{-1.5em}
    \caption{
    Our framework consists of two main stages: static scene reconstruction and dynamic scene generation. The static stage uses a hybrid 3D Gaussian representation to reconstruct scenes from multiple trajectories, with a multi-node design and trajectory embeddings to decouple scene components and environmental conditions. 
    Given a reference video and a target view, the dynamic stage combines the reconstructed static scene with diffusion-based generation to synthesize view-consistent dynamic scene.
    }
    \vspace{-1.5em}
    \label{fig:pipeline}
\end{figure*}

Our goal is to enable high-fidelity, controllable simulation of complex real-world driving scenarios, accurately capturing both static scenes and dynamic objects under diverse conditions. To achieve this, our framework combines static scene reconstruction with dynamic scene generation as shown in \autoref{fig:pipeline}. We use multi-view driving sequences to build a 3DGS-based static model that captures appearance variations and supports high-quality, flexible rendering. Leveraging these static priors, we generate dynamic scenes from arbitrary viewpoints, simulating diverse and realistic traffic scenarios in a highly controllable manner.


\subsection{Static Scene Reconstruction}\label{sec:static}

Given a set of multi-view driving images $I = \{I_{ij}\}$, where $i$ indexes frames within each traverse and $j$  indexes traverses, our goal is to reconstruct the static environment using a 3D Gaussian Splatting (3DGS) framework. In vanilla 3DGS, the scene is represented by a set of $N$ 3D Gaussians, each parameterized by its position $\mathbf{x}$, rotation $\mathbf{q}$, scale $\mathbf{s}$, opacity $\alpha$, and a view-dependent color $\mathbf{c}(\mathbf{d})$. The color is typically modeled as a spherical harmonics (SH) expansion, representing view-dependent effects as a weighted sum of SH basis functions evaluated at the viewing direction $\mathbf{d}$. For rendering, each Gaussian is projected onto the image plane and $\alpha$-blended. The final pixel color $\mathbf{C}$ is computed by back-to-front compositing:
\begin{equation}\label{eq:rasterization}
    \mathbf{C} = \sum_{k=1}^{N} \mathbf{c}_k(\mathbf{d}) \, \alpha_k \prod_{l=1}^{k-1} (1 - \alpha_l),
\end{equation}
where $N$ is the total number of Gaussians in the scene and $\alpha_k$ is the opacity of the $k$-th Gaussian.

However, vanilla 3DGS assumes relatively uniform and static scene appearance, which is often insufficient for modeling the diverse and complex conditions encountered in real-world autonomous driving scenarios. In particular, such scenes may exhibit appearance variations across traverses, and mutable content in geometry (e.g., building ornaments, vegetation, or billboards). These challenges motivate us to extend the standard framework with more expressive and adaptable Gaussian representations tailored to different scene components.
As shown in ~\autoref{fig:pipeline}, we organize the scene into three types of nodes: \textbf{sky nodes}, \textbf{ground nodes}, and \textbf{background nodes}.

\subsubsection{Sky and Ground Nodes}
Sky and ground regions typically exhibit coherent color patterns within each traversal, but their overall appearance can vary substantially across different traversals due to changes in weather and lighting conditions. To address this, we replace the traditional spherical harmonics (SH) coefficients for view-dependent color with a set of learnable feature codes~\cite{yang2023learning,tonderski2024neurad,hess2025splatad}, resulting in a more flexible and expressive representation, which we refer to as Code-Gaussians. Specifically, each Code-Gaussian is parameterized as $\mathcal{G}_{\text{code}} = \{\mathbf{x}, \mathbf{q}, \mathbf{s}, \alpha, \mathbf{f}\}$, where $\mathbf{x}$ denotes position, $\mathbf{q}$ rotation, $\mathbf{s}$ scale, $\alpha$ opacity, and $\mathbf{f}$ is a learnable code.
To capture appearance variations across traversals, we introduce an appearance latent vector $\mathbf{z}_j = \mathrm{Emb}(j)$ by embedding the traversal ID $j$, which encodes traversal-specific appearance factors. The color of each Gaussian is then predicted as a function of the view direction $\mathbf{d}$, conditioned on the appearance latent $\mathbf{z}_j$ and the Gaussian’s feature $\mathbf{f}$:
\begin{equation}
    \mathbf{c}(\mathbf{d}) = \mathrm{MLP}(\mathbf{d} \mid \mathbf{z}_j, \mathbf{f}).
\end{equation}
This design enables the model to capture both smooth color transitions within a single traversal and significant global appearance variations across different traversals, resulting in a more adaptive and expressive representation for diverse autonomous driving scenarios.

\subsubsection{Background Nodes}
Background nodes refer to the set of Gaussian points excluding those assigned to the sky and ground. These background regions often contain more intricate and fine-grained geometric structures, and are subject to both appearance and geometric variations across different traversals. To effectively model such complexity, we adopt an anchor-controlled ScaffoldGS~\cite{lu2024scaffold} representation for background nodes, parameterized by position $\mathbf{x}$ and a learnable code $\mathbf{f}$. Instead of fitting a redundant set of Gaussians to every training view, ScaffoldGS leverages anchor points to distribute local 3D Gaussians, resulting in a more compact and geometrically consistent representation. For each anchor point $A$, a scaffold decoder dynamically predicts the attributes of its associated offset Gaussians based on both scene and view-dependent factors. Specifically, given the $j$-th traversal's appearance latent $\mathbf{z}_j$, the anchor's position $\mathbf{x}_A$ and code $\mathbf{f}_A$, the offset feature $\mathbf{f}_k$, and the viewing direction $\mathbf{d}$, the decoder predicts the parameters of each offset Gaussian as:
\begin{equation}
\mathbf{x}_k, \mathbf{q}_k, \mathbf{s}_k, \alpha_k, \mathbf{c}_k = \mathrm{MLP}(\mathbf{z}_j, \mathbf{x}_A, \mathbf{f}_A, \mathbf{f}_k, \mathbf{d}), \quad k \in \mathcal{K}_A,
\end{equation}
where $\mathcal{K}_A$ denotes the set of offsets associated with anchor $A$. This approach effectively reduces redundancy, enhances robustness to view changes, and enables efficient coverage of complex and dynamic scene geometry.

\subsubsection{Loss Functions}
We supervise the model using a combination of RGB photometric loss, depth supervision, and normal consistency loss. The total loss is defined as:
\begin{equation}
\mathcal{L}_{\text{total}} = \mathcal{L}_{\text{rgb}} + \lambda_{\text{depth}}\mathcal{L}_{\text{depth}} + \lambda_{\text{normal}}\mathcal{L}_{\text{normal}},
\end{equation}
where $\lambda_{\text{depth}}$ and $\lambda_{\text{normal}}$ are balancing weights. The loss definitions and implementation details are in supplementary material.

\subsection{Dynamic Scene Generation}
\label{sec:dynamic_generation}
To synthesize realistic and temporally coherent dynamic driving scenes along novel trajectories, we combine the geometric consistency of 3DGS with the generative power of video diffusion models. Although 3DGS reconstructs static environments well, modeling dynamic agents in 3D often introduces artifacts and limits motion flexibility. Video diffusion models, on the other hand, provide strong generative priors but may suffer from cross-view inconsistency and temporal drift. We therefore propose an autoregressive framework that combines a static 3DGS background with a history-conditioned video diffusion model, as shown in~\autoref{fig:pipeline}.

Given a novel target trajectory \(\tau^{tgt}\), our goal is to synthesize the corresponding target multi-view video \(\mathcal{V}^{out}\) by generating dynamic agents that are consistent with both the underlying static scene geometry and the temporal context. To support trajectories of arbitrary length, we formulate dynamic scene generation as an autoregressive process with a sliding temporal window. At each time step \(t\), the model predicts the current target frame \(I_t^{out}\) based on a short sequence of preceding reference frames
\[
\mathcal{V}^{ref}_{t-T:t-1} = \{I_{t-T}^{ref}, \dots, I_{t-1}^{ref}\}.
\]
During training, these reference frames are taken from the ground-truth target sequence using teacher forcing, whereas during inference they are replaced by the model's previously generated outputs. This autoregressive formulation enables long-horizon synthesis while preserving temporal continuity across frames.

Based on the static scene reconstruction described in Section~\ref{sec:static}, we obtain a hybrid Gaussian background representation, denoted as \(F_{BG}\). For each frame along the target trajectory, the static renderer takes the target view direction \(d_t\) and an appearance latent \(L\) as input, and produces the background image
\[
I_t^{bg} = F_{BG}(d_t, L).
\]
We denote the rendered static multi-view background sequence as
\[
\mathcal{V}^{bg} = \{I_1^{bg}, I_2^{bg}, \dots \},
\]
where each \(I_t^{bg}\) contains the rendered background images for all target camera views at time step \(t\). In this way, the generated dynamic content is anchored to a background that remains geometrically consistent with the reconstructed scene while preserving the desired appearance attributes.

However, naive autoregressive generation tends to accumulate errors over time and may gradually drift away from the underlying scene geometry. To alleviate this issue, we augment each prediction step with both autoregressive visual references and auxiliary scene conditions. Specifically, at time step \(t\), the model takes as input a sequence of preceding reference frames
\[
\mathcal{V}^{ref}_{t-T:t-1} = \{I_{t-T}^{ref}, \dots, I_{t-1}^{ref}\},
\]
which provides short-term temporal context for preserving motion continuity and appearance consistency. During training, these reference frames are taken from the ground-truth target sequence using teacher forcing, whereas during inference they are replaced by the model's previously generated outputs.

In addition to the rendered background images and autoregressive reference frames, we introduce several auxiliary conditioning signals for the current prediction. First, we extract 3D layout cues from the reference frames and the current target frame, denoted as \(B_t^{ref}\) and \(B_t^{tgt}\), to provide object-level spatial guidance. Second, we construct a camera raymap \(R_t\) to encode the target camera configuration. Finally, to inject high-level semantic context, we summarize the reference frames into a textual prompt \(P_t\) using a vision-language model. We denote these auxiliary conditions as
\[
\mathcal{C}_t = \left( B_t^{ref}, B_t^{tgt}, R_t, P_t \right).
\]
Together, the autoregressive reference frames and the auxiliary conditioning signals guide the model to generate dynamic scenes that are both temporally coherent and geometrically aligned with the reconstructed static environment.

For training, we construct a sample at each time step \(t\) using the ground-truth target frame and its corresponding inputs, as shown in~\autoref{fig:pipeline}. We first encode the ground-truth target multi-view frame at time step \(t\) into a clean latent representation \(x_0\) using a VAE. We then sample a diffusion timestep \(s\) and add Gaussian noise \(\epsilon \sim \mathcal{N}(0,\mathbf{I})\) to obtain the noisy latent \(x_s\). The reference frames in \(\mathcal{V}^{ref}_{t-T:t-1}\) are encoded by the same VAE into reference latent features \(z^{ref}\). In parallel, the rendered background image \(I_t^{bg}\) is processed by a dedicated image adapter, implemented as a lightweight multi-scale network with residual blocks and zero-initialized convolutions~\cite{mou2024t2i}, to extract background features \(f_t^{bg}\). The noisy latent \(x_s\), the reference latent features \(z^{ref}\), and the adapted background features \(f_t^{bg}\) are then fused together and fed into the denoising transformer, together with the auxiliary conditional inputs $C_t$. The denoising network \(\Phi\) is thus trained to predict the added noise conditioned on both latent-space references and adapter-based scene cues:
\[
\mathcal{L}_{\mathrm{diff}}
=
\mathbb{E}_{x_0,\epsilon,s}
\left[
\left\|
\epsilon - \Phi(x_s, s, z^{ref}, f_t^{bg}, C_t)
\right\|_2^2
\right].
\]

At inference time, for each target frame, we first render the background image \(I_t^{bg}\) from \(F_{BG}\) using the target view direction and the specified appearance latent. We then form the reference sequence \(\mathcal{V}^{ref}_{t-T:t-1}\) from the previously generated frames, encode it into latent features, and concatenate these features with the current noisy latent. The resulting latent input, together with the auxiliary conditions \(\mathcal{C}_t\), is fed into the diffusion model to predict the current frame \(I_t^{out}\). For the first \(T\) steps, we initialize the autoregressive process using given seed frames; afterwards, the reference window is updated using the model's own predictions. This design decouples static scene rendering from dynamic content synthesis: the static 3DGS model guarantees cross-view geometric consistency, while the diffusion model provides flexible and temporally coherent generation of dynamic objects.
\begin{figure}[htp]
    \centering
    \vspace{-1.5em}
    \includegraphics[width=0.9\linewidth]{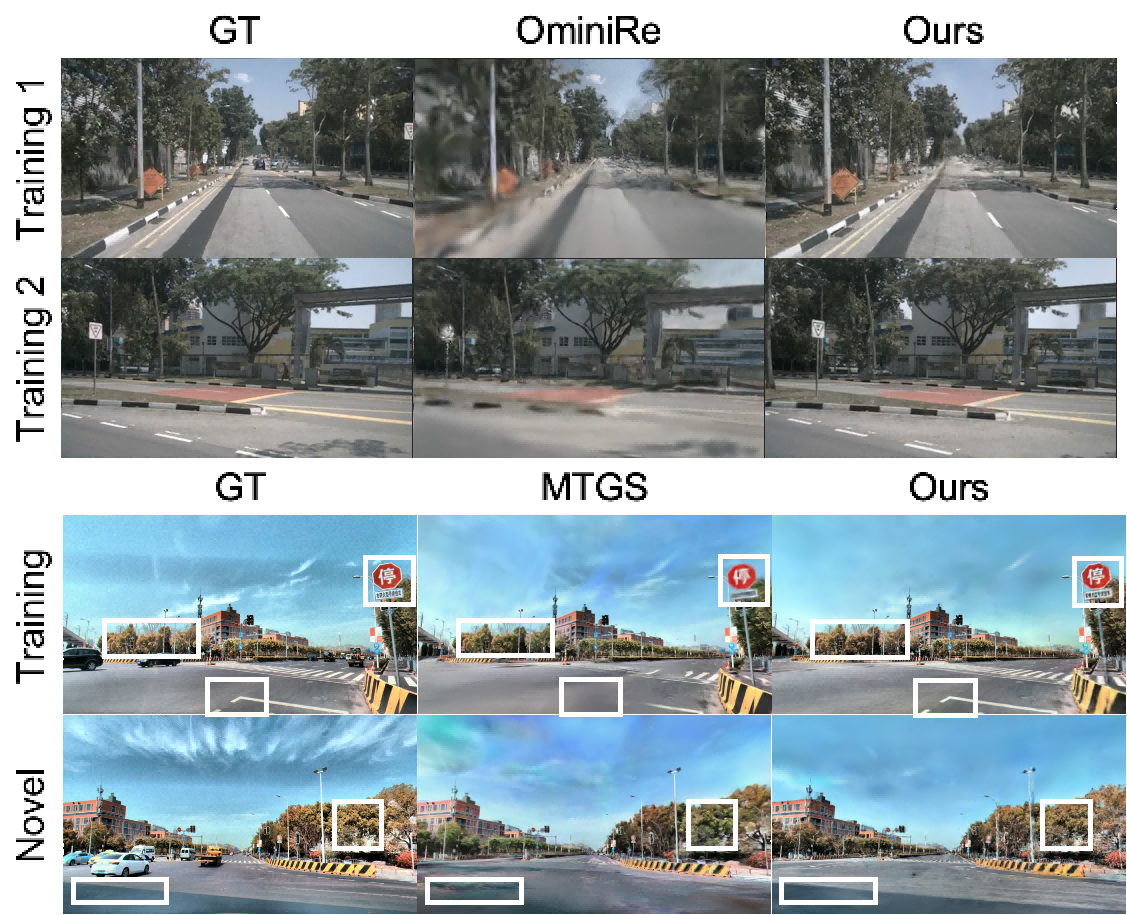}
    \vspace{-1em}
    \caption{
    We compare our static scene reconstruction module with OmniRe~\cite{chen2025omnire} (single-traversal) and MTGS~\cite{li2025mtgs} (multi-traversal) on nuScenes~\cite{caesar2020nuscenes} and the MIRROR dataset.
    }
    \vspace{-2em}
    \label{fig:recon}
\end{figure}

\begin{table*}[htp]
    \centering
    \caption{Quantitative Comparison for Static Scene Reconstruction. Best results are in \textbf{bold}.}
    \vspace{-0.5em}
    \resizebox{0.8\linewidth}{!}{
    \begin{tabular}{l|ccccc ccc}
        \toprule
        \multirow{2}{*}{Dataset} & \multirow{2}{*}{Single/Multi} & \multirow{2}{*}{Method}
        & \multicolumn{3}{c}{Training Traversal}
        & \multicolumn{3}{c}{Novel Traversal} \\
        \cmidrule(lr){4-6} \cmidrule(lr){7-9}
        & & & PSNR $\uparrow$ & SSIM $\uparrow$ & LPIPS $\downarrow$
          & $\text{PSNR}^*$ $\uparrow$ & SSIM $\uparrow$ & LPIPS $\downarrow$ \\
        \midrule
        \multirow{2}{*}{nuScenes~\cite{caesar2020nuscenes}}
            & \multirow{2}{*}{Single} & OmniRe~\cite{chen2025omnire}    
            & 27.70 &0.826 &0.179
            &   --   &   --   &   --   \\
            &                         & Ours 
            & {\bf 30.39} &{\bf 0.886}  &{\bf 0.147}     
            &   --   &   --   &   --   \\
        \midrule
        \multirow{2}{*}{MIRROR}
            & \multirow{2}{*}{Multi}  & MTGS~\cite{li2025mtgs}   
            &   21.40      &   0.693     &  0.381       
            &  16.072    & 0.583        &    0.486     \\
            &                         & Ours  
            & {\bf 22.82}     & {\bf 0.753}        &         {\bf 0.339}
            &  {\bf 17.734}       & {\bf 0.590}        &     {\bf 0.403}    \\
        \midrule
        \multirow{2}{*}{nuPlan~\cite{caesar2021nuplan}}
            &   \multirow{2}{*}{Multi}                      & MTGS~\cite{li2025mtgs}  &  28.04       &   0.848      & 0.192 &   21.65     &    {\bf 0.628}     &   0.265      
            \\
            &                         & Ours  &   {\bf 29.17}      &   {\bf 0.873}      &   {\bf 0.129}      &  {\bf 22.11}       &    0.606      &  {\bf 0.219}  \\
        \bottomrule
    \end{tabular}
    }
    \label{tab:static_comparison}
    \vspace{-0.8em}
\end{table*}
\begin{figure*}[htp]
    \centering
    \includegraphics[width=0.85\textwidth]{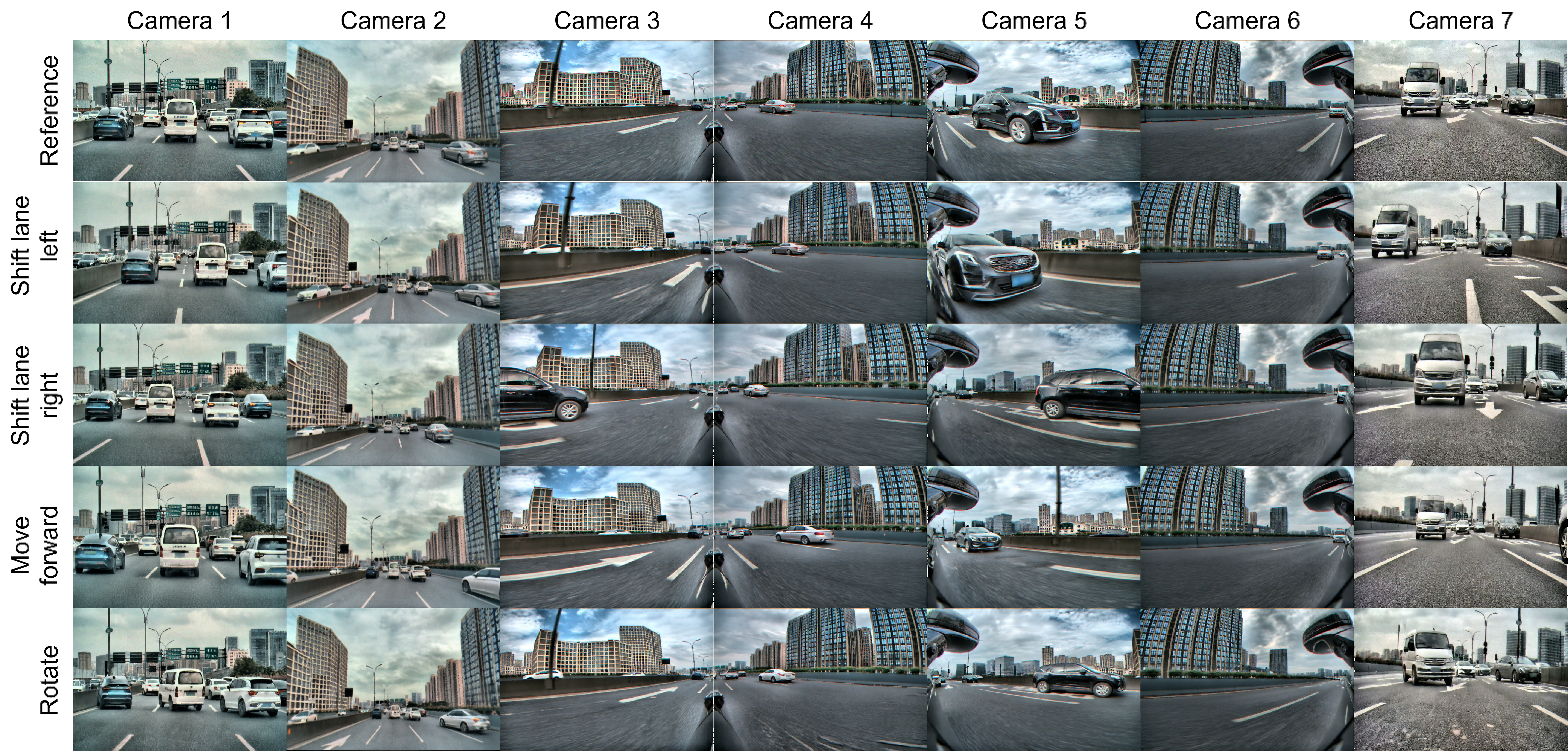}
    \vspace{-1em}
    \caption{
Novel ego-trajectory synthesis results.
We present reference views (top) and synthesized target views under diverse driving behaviors, including lane shifts, forward motion, and dynamic agent editing.
    }
    \vspace{-1em}
    \label{fig:supp_gallery}
\end{figure*}

\section{Experiments}
\noindent\textbf{Datasets.}
We evaluate our method on three datasets: nuPlan~\cite{caesar2021nuplan}, nuScenes~\cite{caesar2020nuscenes}, and our MIRROR dataset.

\noindent\textbf{Baselines and Metrics.}
We compare our method with 4 representative baselines: \textbf{(a)} OmniRe~\cite{chen2025omnire}, a driving scene reconstruction method; \textbf{(b)} MTGS~\cite{li2025mtgs}, a multi-trajectory reconstruction method; \textbf{(c)} DriveEditor~\cite{liang2024driveeditor}, a vehicle editing method; and \textbf{(d)} MagicDrive-V2~\cite{gao2025magicdrive-v2}, a controllable driving video generation framework.
Evaluation is performed from three perspectives: static scene reconstruction, dynamic scene generation, and closed-loop simulation. For reconstruction quality, we report PSNR, SSIM, and LPIPS. For controlled dynamic intervention, we use FID to assess generation fidelity and consistency. Closed-loop performance is evaluated qualitatively.

\noindent\textbf{Implementation Details.}
Static scene reconstruction is optimized with Adam for 90,000 iterations, and the Gaussian renderer achieves 21 FPS at \(960 \times 540\). The dynamic generation model is trained using frame pairs \((\mathcal{V}^{ref}_{t-T:t-1}, I^{out}_t)\), where \(T=5\). All compared methods use their official codebases with default settings; MTGS is trained for 90,000 iterations on MIRROR for fair comparison. Additional details are provided in the supplementary material.

\subsection{Static Background Reconstruction}
We evaluate our method against OmniRe~\cite{chen2025omnire} on nuScenes~\cite{caesar2020nuscenes} for single-traversal reconstruction and compare with MTGS~\cite{li2025mtgs} for multi-traversal reconstruction on nuScenes, nuPlan~\cite{caesar2021nuplan}, and MIRROR. Dynamic objects are masked during evaluation to focus on static scene quality.
For nuPlan, we follow MTGS~\cite{li2025mtgs} and use six road blocks with LiDAR initialization and 30k training iterations; for nuScenes, we use the mini split; and for MIRROR, we use 4 traversals for training and 1 held-out traversal for novel-view evaluation. To ensure a fair comparison, all methods are initialized with the same Gaussian initialization.
As shown in \autoref{tab:static_comparison}, our method outperforms OmniRe in reconstruction quality. Qualitative results in \autoref{fig:recon} show that our reconstructions are noticeably sharper than OmniRe across all scenes.
Compared with MTGS, our method achieves better performance on training views and stronger generalization to novel views, particularly on the more challenging MIRROR dataset, which features diverse lighting conditions, wide roads, and frequent lane changes. \autoref{fig:recon} further shows that our approach recovers finer details in lane markings and traffic signs, and better captures seasonal foliage variations. These results validate the robustness of our 3DGS reconstruction in complex environments.


\subsection{Dynamic Scene Generation}
Our primary dynamic result is novel ego-trajectory synthesis, generating temporally coherent target views under unseen ego motions while preserving geometry and dynamic plausibility. As shown in \autoref{fig:supp_gallery}, our method maintains consistent rendering across cameras even under large viewpoint changes, with robust appearance in side views despite significant cross-camera shifts.

To evaluate controllability, we adopt a controlled vehicle translation benchmark and compare with DriveEditor~\cite{liang2024driveeditor} on both nuScenes~\cite{caesar2020nuscenes} and MIRROR, using 1 scene from nuScenes and 4 scenes from MIRROR. The evaluation applies parallel offsets to dynamic vehicles and measures FID to assess generation fidelity and consistency.

As shown in \autoref{tab:gen_fidelity}, our method maintains superior consistency across different offsets, particularly under extreme displacements (\(Y=\pm 3\) meters).
Additional qualitative comparisons are provided in the supplementary material. These results show that our approach produces vehicles with more reasonable scale and more seamless background integration while preserving stronger geometric and appearance alignment with the reference image, enabling more accurate scaling and placement. 
In contrast, DriveEditor exhibits noticeable scale inconsistencies. These advantages stem from our conditioning mechanism, which effectively enforces photometric and geometric consistency throughout the generation process.

\begin{table}[htp]
\centering
\caption{
    FID ($\downarrow$) comparison for controlled vehicle translation under different horizontal offsets $Y$. Lower FID indicates better generation fidelity. Better results are in bold.
}
\vspace{-1em}
\renewcommand{\arraystretch}{1.15}
\setlength{\tabcolsep}{8pt}
\resizebox{\linewidth}{!}{
\begin{tabular}{l l c c c c c}
\toprule
\textbf{Dataset} & \textbf{Method} & $Y=-2$ & $Y=-1$ & $Y=0$ & $Y=1$ & $Y=2$ \\
\midrule
\multirow{2}{*}{nuScenes~\cite{caesar2020nuscenes}} 
& DriveEditor~\cite{liang2024driveeditor} & 75.01 & 55.96 & 47.14 & 61.06 & 72.48 \\
& \textbf{Ours} & \textbf{25.34} & \textbf{21.65} & \textbf{19.80} & \textbf{22.01} & \textbf{37.79} \\
\midrule
 & & $Y=-3$ & $Y=-1.5$ & $Y=0$ & $Y=1.5$ & $Y=3$ \\
\midrule
\multirow{2}{*}{MIRROR} 
& DriveEditor~\cite{liang2024driveeditor} & 30.26 & 26.03 & 23.51 & 25.74 & 27.22 \\
& \textbf{Ours} & \textbf{18.94} & \textbf{18.13} & \textbf{17.98} & \textbf{18.40} & \textbf{23.17} \\
\bottomrule
\end{tabular}
}
\vspace{-2em}
\label{tab:gen_fidelity}
\end{table}

\subsection{Closed-loop Simulator}
We compare our proposed world simulator with both MagicDrive-V2~\cite{gao2025magicdrive-v2} and OmniRe~\cite{chen2025omnire} on nuScenes~\cite{caesar2020nuscenes}. MagicDrive-V2 is a purely generative approach that takes a BEV map, foreground bounding boxes, and the ego trajectory as input to synthesize multi-view driving videos, while OmniRe is a reconstruction-based Gaussian rendering method. 
Since these approaches differ substantially in task formulation and supervision, we focus on qualitative evaluation.

Qualitative comparisons are provided in the supplementary material. The results show that our method achieves stronger geometric consistency in object structures and scene backgrounds than MagicDrive-V2. 
Compared with OmniRe, which manipulates vehicles by editing their 3DGS representations, our framework produces more realistic results without artifacts, highlighting the advantage of our framework in high-fidelity dynamic simulation. Notably, for the dynamic generation results on nuScenes, we use our model in an inference-only manner without additional training, demonstrating encouraging cross-dataset generalization.

\subsection{Limitations}
Although our method achieves promising dynamic generation quality, its inference efficiency is still limited by the multi-step sampling process of diffusion models. In the current implementation, generating each 7-view frame requires 10 sampling steps. An important direction for future work is to distill the model to fewer inference steps for faster generation.
\section{Conclusion}
In this work, we present HybridWorldSim, a simulation framework that combines 3D Gaussian-based static scene reconstruction with generative modeling for dynamic scene synthesis. Leveraging multi-traversal data, especially our newly introduced MIRROR dataset, our approach enables high-fidelity and controllable simulation of complex real-world driving scenarios. Experiments on public benchmarks and MIRROR demonstrate consistent improvements over existing baselines in novel view synthesis and scene editing. These results suggest that HybridWorldSim provides a practical and extensible foundation for autonomous driving research and evaluation.

\begin{acks}
This work was conducted during the internships of Yingwenqi Jiang and Xiang Feng at XPeng Inc. 
\end{acks}

{
    \bibliographystyle{ACM-Reference-Format}
    \balance
    \bibliography{main}
}

\clearpage
\setcounter{page}{1}

\appendix

To provide a more comprehensive understanding of our work, this supplementary material presents additional details and results beyond those included in the main paper. Along with expanded dataset statistics and qualitative examples, we provide clarifications on several implementation details that were only briefly discussed previously, covering dataset specifics, the design of two key modules, and further experimental results. This supplementary information is intended to better demonstrate the diversity and realism of our dataset, the robustness of our approach, and its practical relevance for real-world autonomous driving applications.
To support reproducible research, we will release training checkpoints, inference code, and the full MIRROR dataset.

\section{Details on MIRROR Dataset}
\label{supp:dataset}

\subsection{Scene Coverage} 
To ensure the broad applicability and generalization capability of our approach, we construct the dataset to cover a diverse set of driving environments. As illustrated by the BEV (Bird's Eye View) maps in \autoref{fig:bev_maps}, our dataset encompasses four representative scene types: urban downtown, highway, rural roads, and campus areas. Each scene is carefully selected to capture unique layout characteristics, traffic conditions, and visual appearances, ranging from densely populated city intersections to open highways, rural landscapes, and structured campus roads. This variety enables comprehensive evaluation across different navigation challenges, road geometries, and traffic participant distributions. 

\begin{table}[htp]
\centering
\caption{
Dataset Scene Diversity. Our dataset contains ten scenes organized into four categories: urban, highway, rural, and campus. Each scene is visualized through bird's-eye view (BEV) maps and semantic segmentation, highlighting the diversity of driving environments—from complex urban intersections and structured highway networks to open rural roads and distinctive campus layouts.
}
\begin{tabular}{|l|cc|}
\hline
Category                 & \multicolumn{1}{c}{BEV} & \multicolumn{1}{c|}{Semantic} 
\\ \hline
\multirow{3}{*}{Urban}   &     
\begin{subfigure}[b]{0.15\textwidth}
    \centering
    \includegraphics[width=1\linewidth]{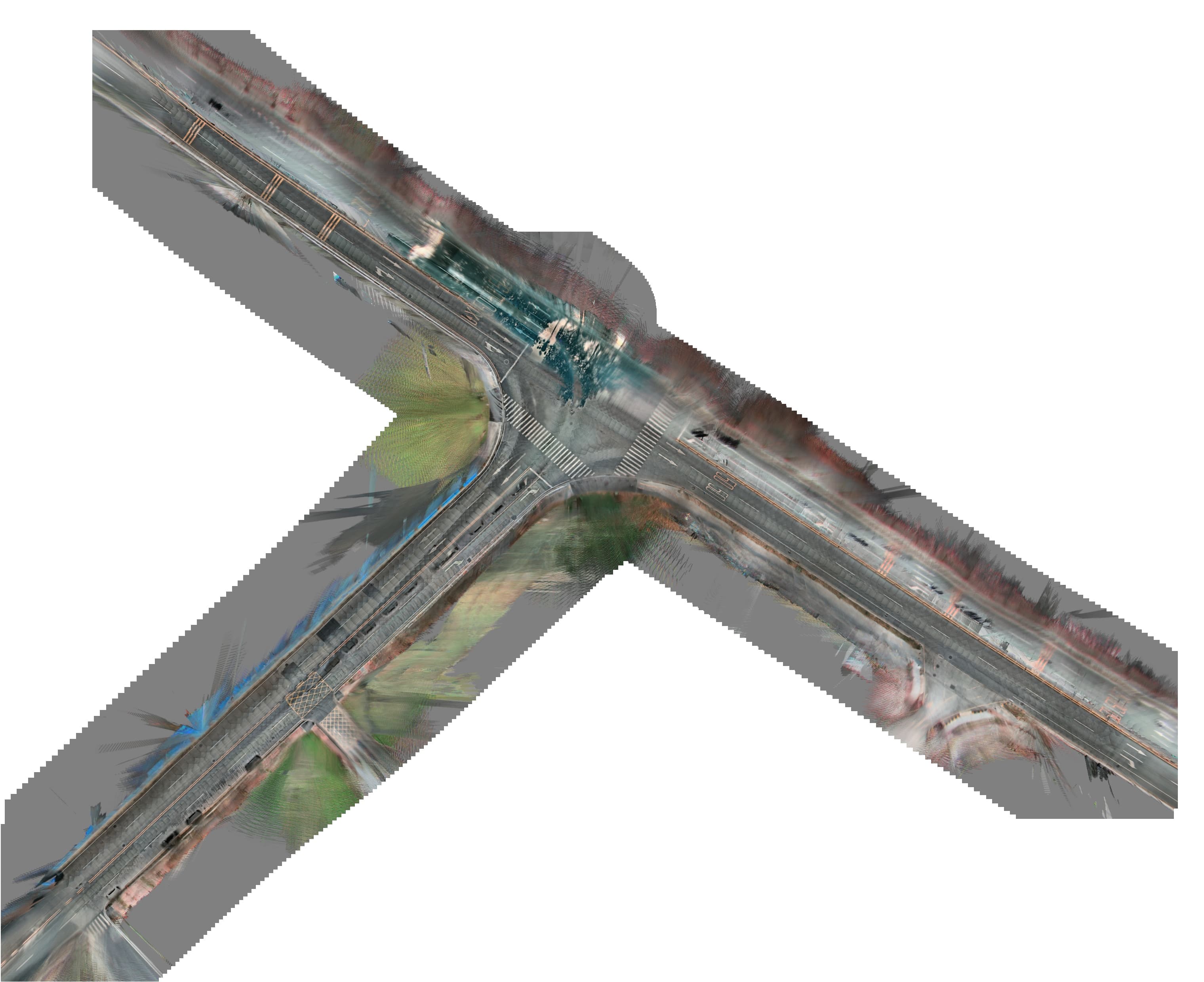}
\end{subfigure}                    
&
 \begin{subfigure}[b]{0.15\textwidth}
    \centering
    \includegraphics[width=1\linewidth]{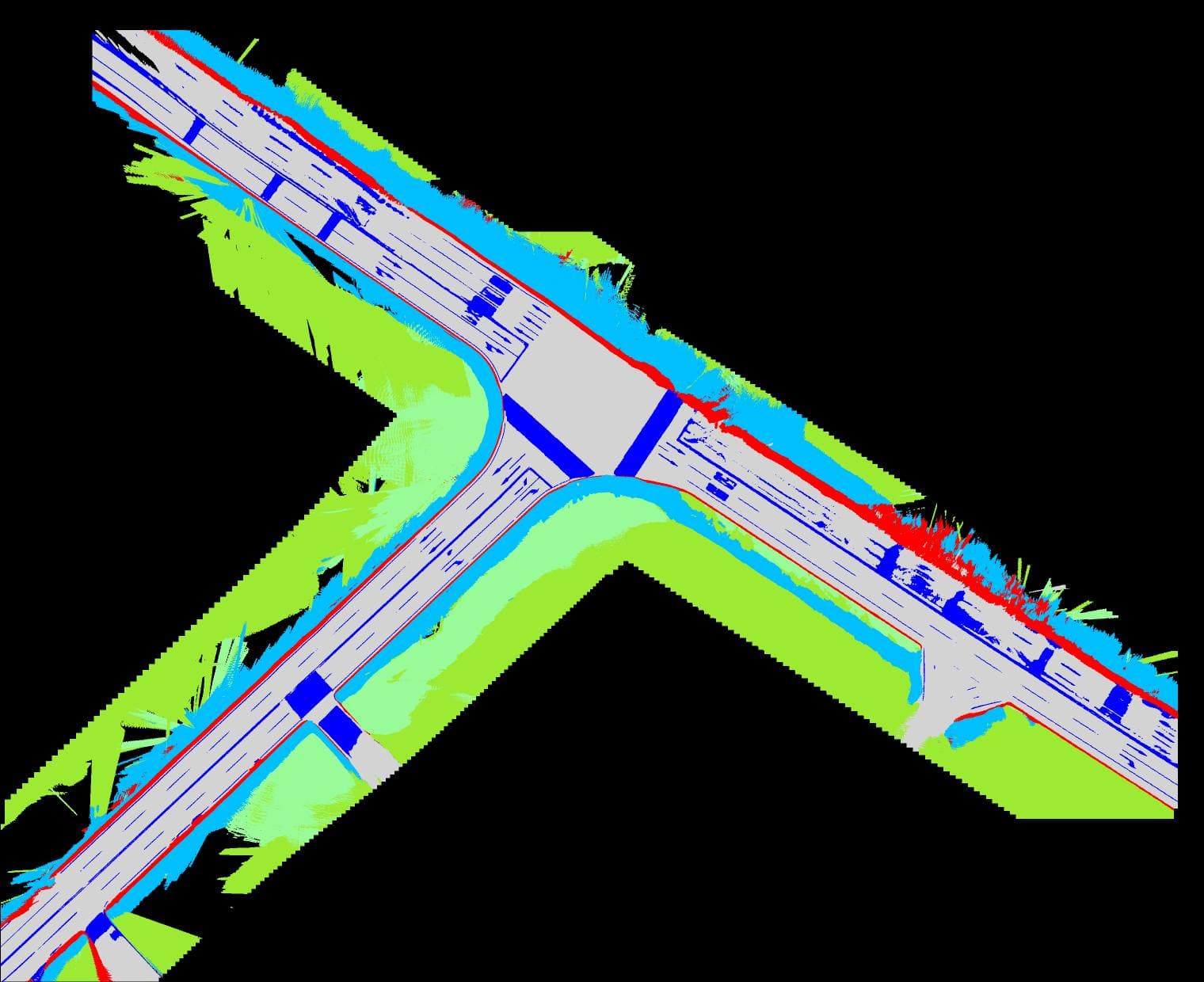}
\end{subfigure}
\\
&
\begin{subfigure}[b]{0.15\textwidth}
    \centering
    \includegraphics[width=1\linewidth]{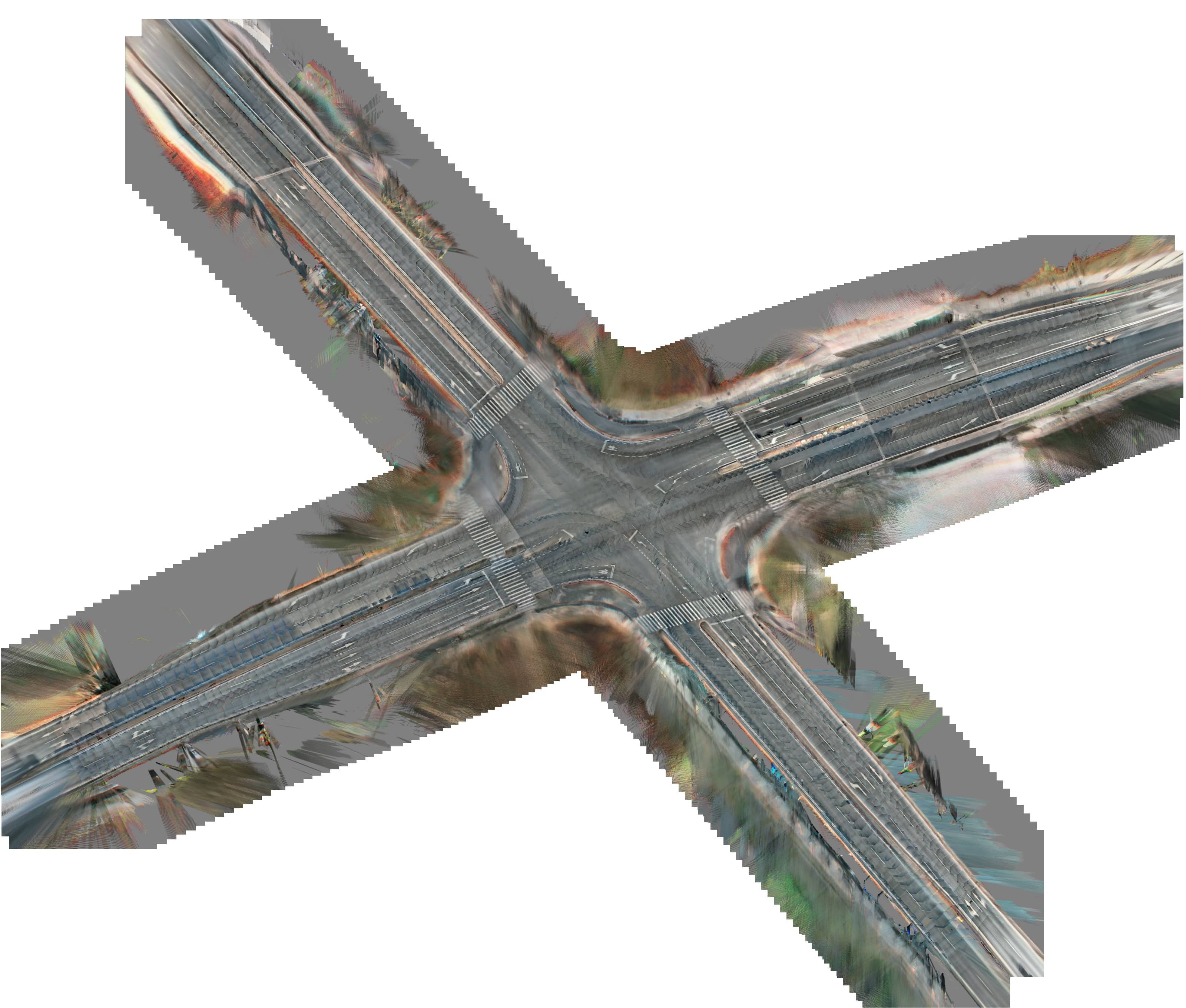}
\end{subfigure}
& 
 \begin{subfigure}[b]{0.15\textwidth}
    \centering
    \includegraphics[width=1\linewidth]{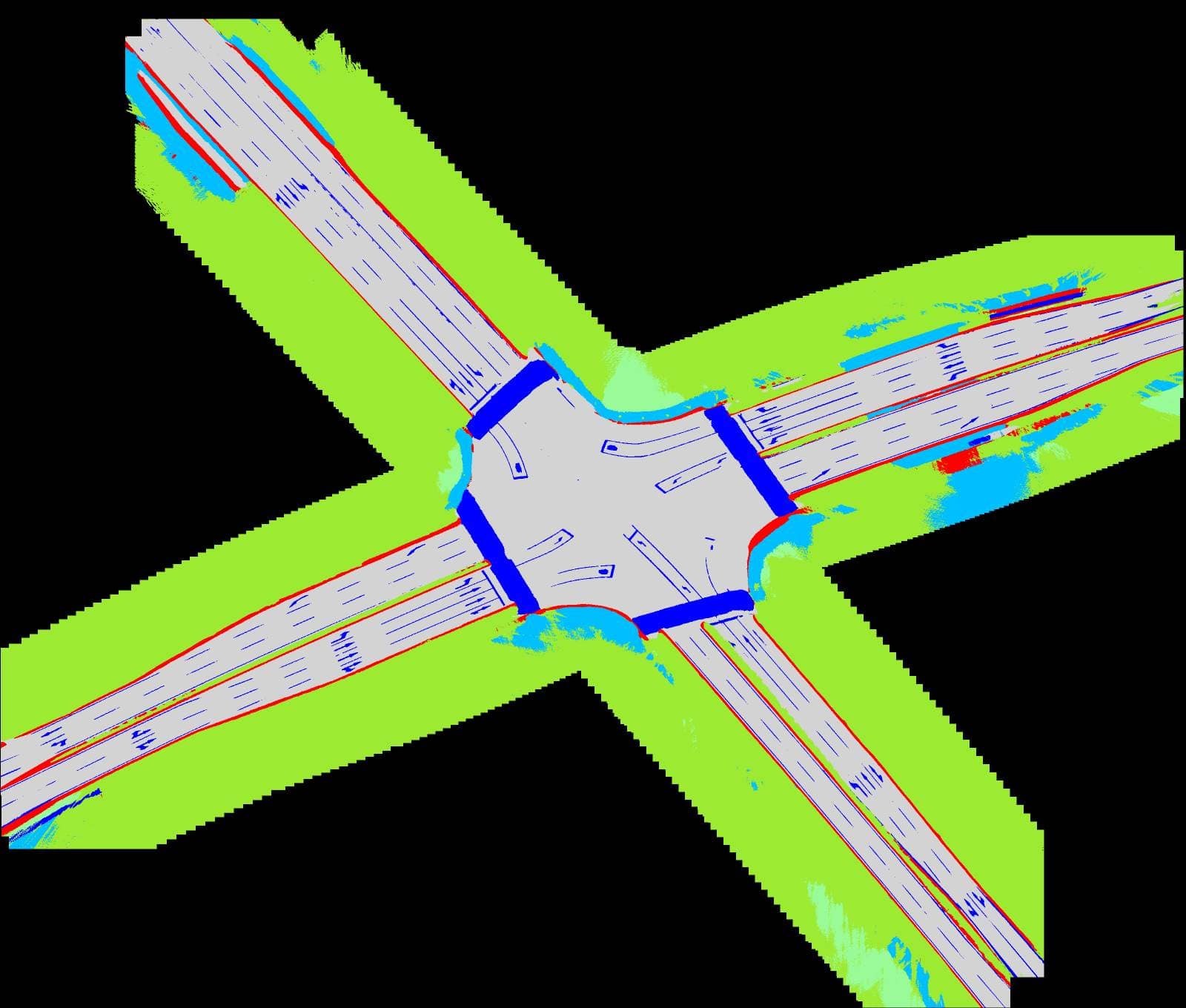}
\end{subfigure}
\\
& 
 \begin{subfigure}[b]{0.15\textwidth}
    \centering
    \includegraphics[width=1\linewidth]{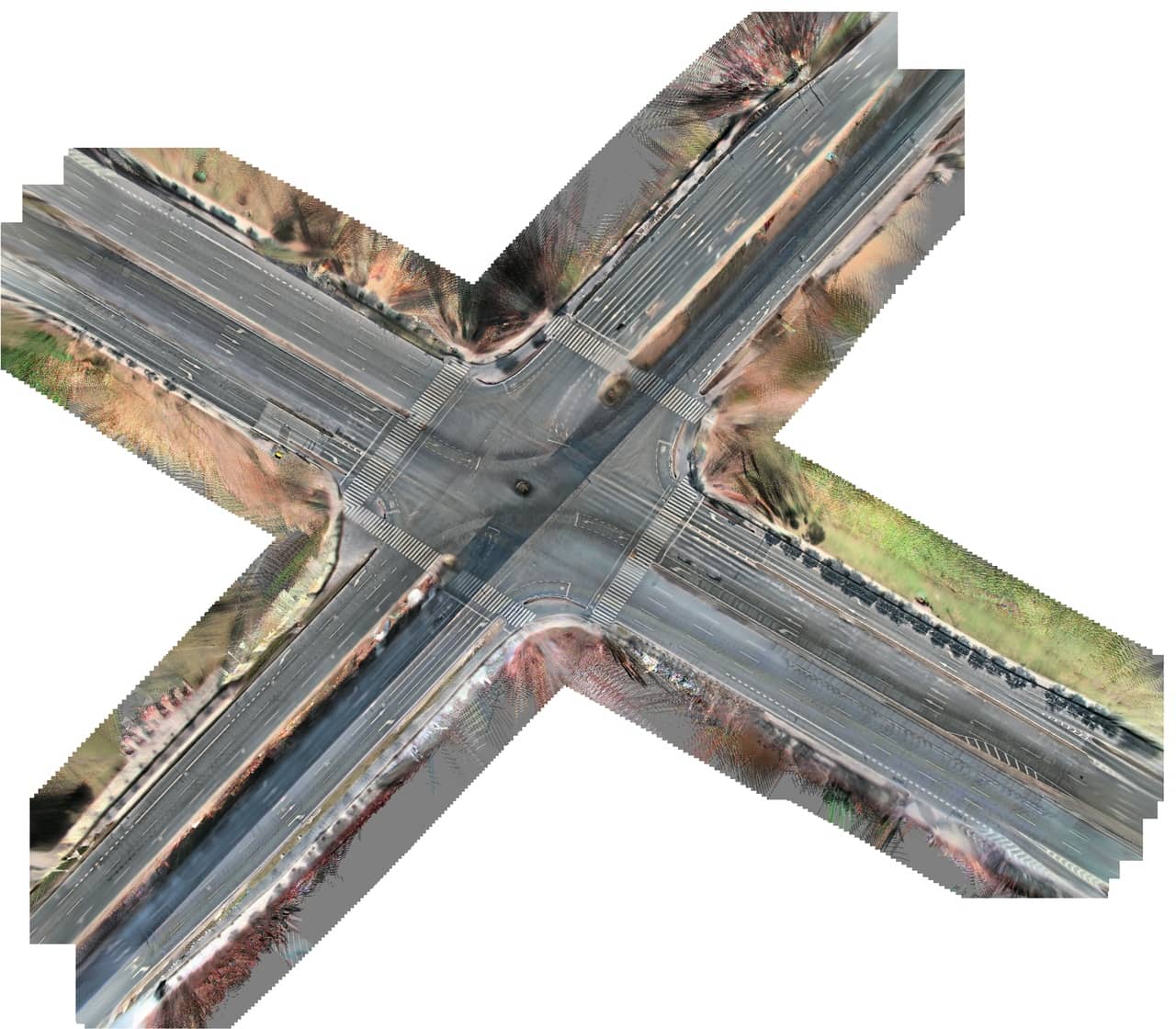}
\end{subfigure}
&
 \begin{subfigure}[b]{0.15\textwidth}
    \centering
    \includegraphics[width=1\linewidth]{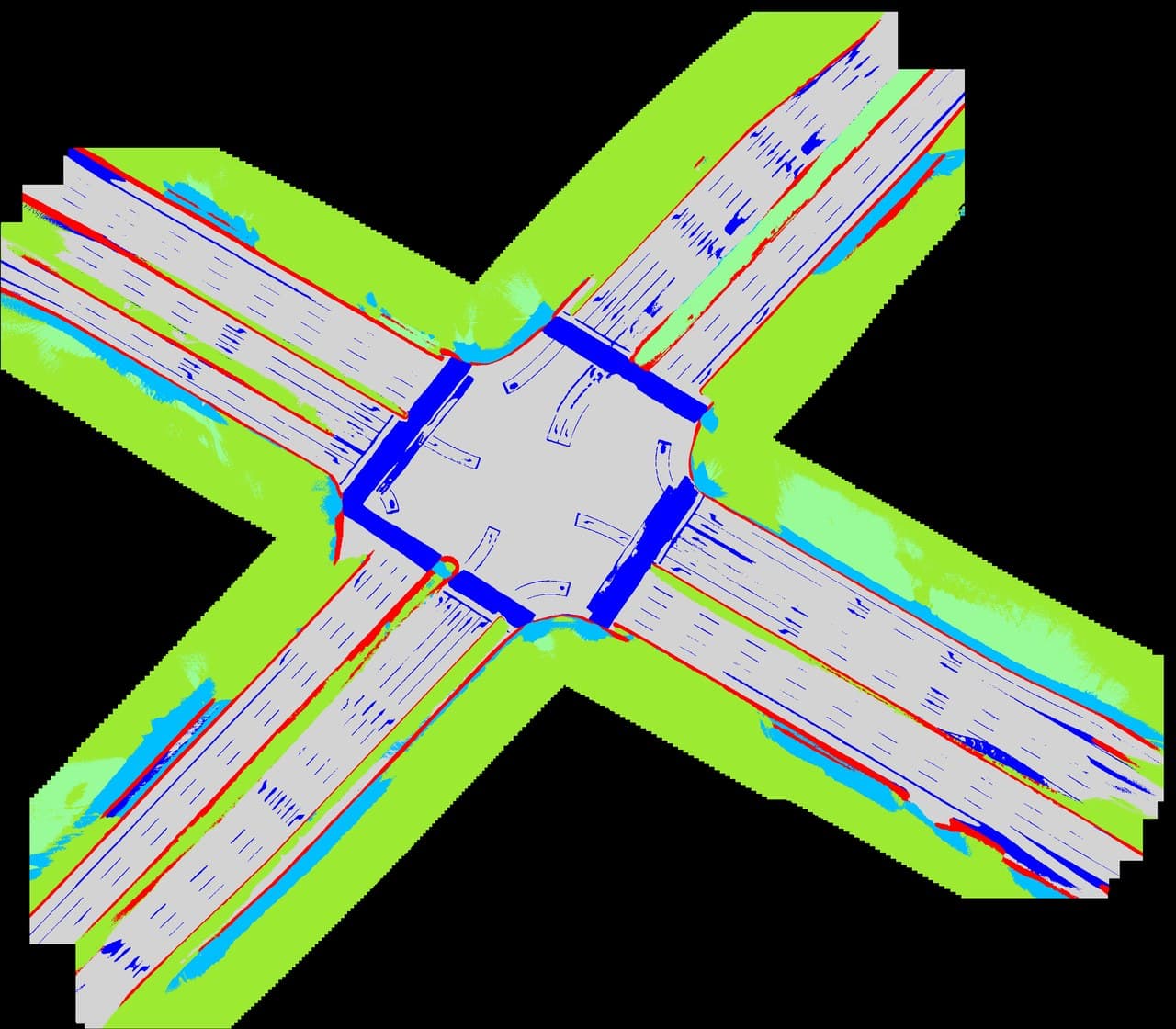}
\end{subfigure}
\\ \hline
\multirow{2}{*}{Highway} &
 \begin{subfigure}[b]{0.15\textwidth}
    \centering
    \includegraphics[width=1\linewidth]{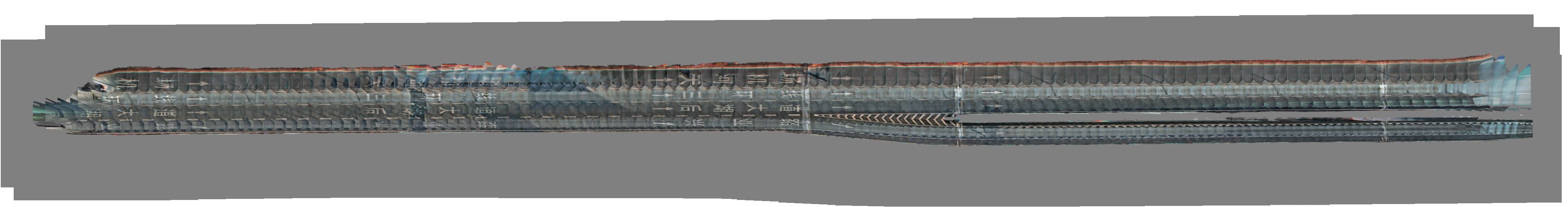}
\end{subfigure}
&
 \begin{subfigure}[b]{0.15\textwidth}
    \centering
    \includegraphics[width=1\linewidth]{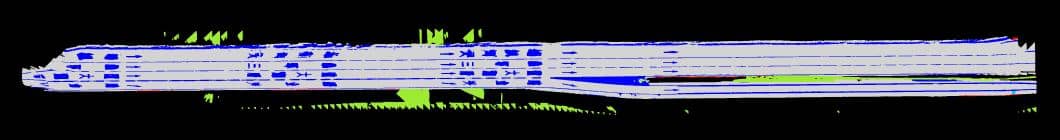}
\end{subfigure}
\\
&
 \begin{subfigure}[b]{0.15\textwidth}
    \centering
    \includegraphics[width=1\linewidth]{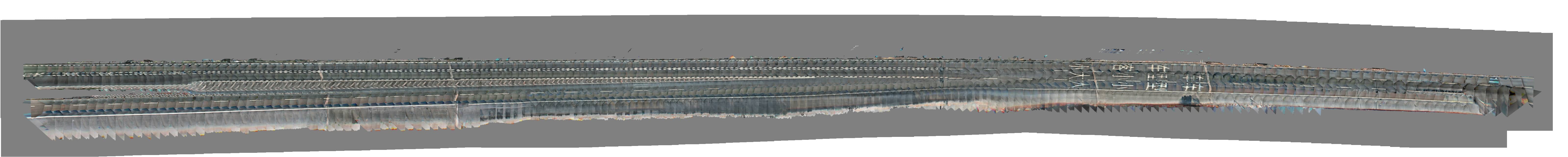}
\end{subfigure}
&
 \begin{subfigure}[b]{0.15\textwidth}
    \centering
    \includegraphics[width=1\linewidth]{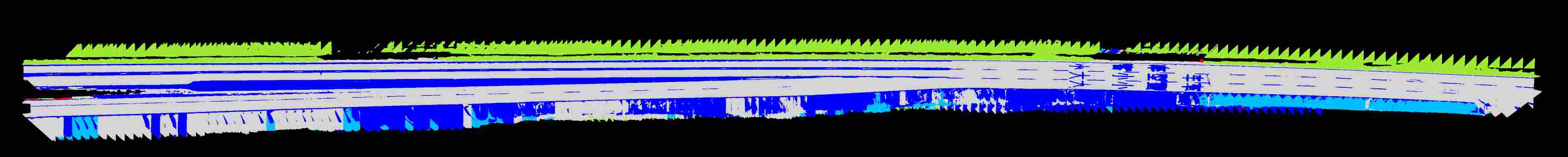}
\end{subfigure}
\\ \hline
\multirow{4}{*}{Campus}  &
\begin{subfigure}[b]{0.15\textwidth}
    \centering
    \includegraphics[width=1\linewidth]{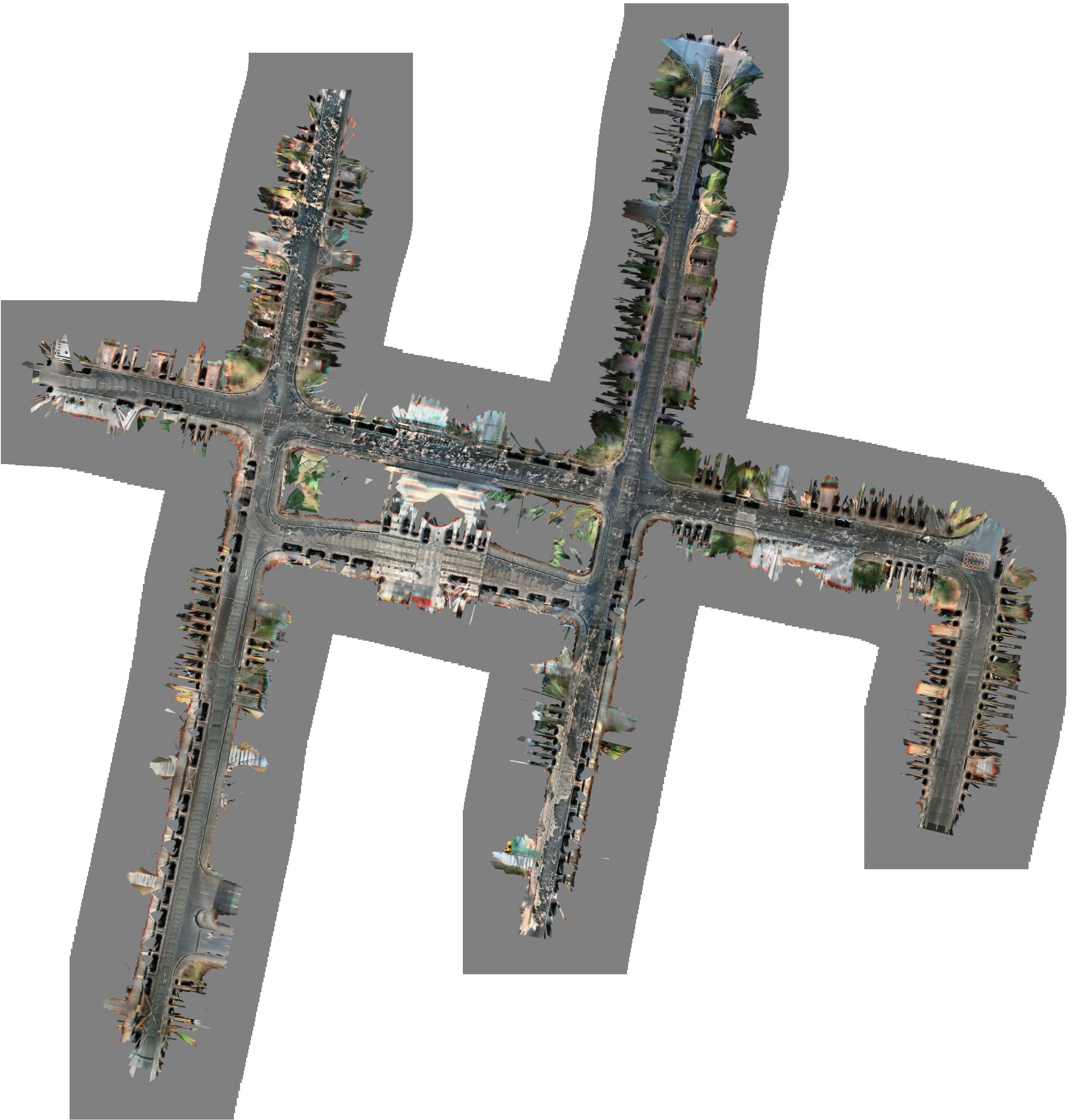}
\end{subfigure}
&
\begin{subfigure}[b]{0.15\textwidth}
    \centering
    \includegraphics[width=1\linewidth]{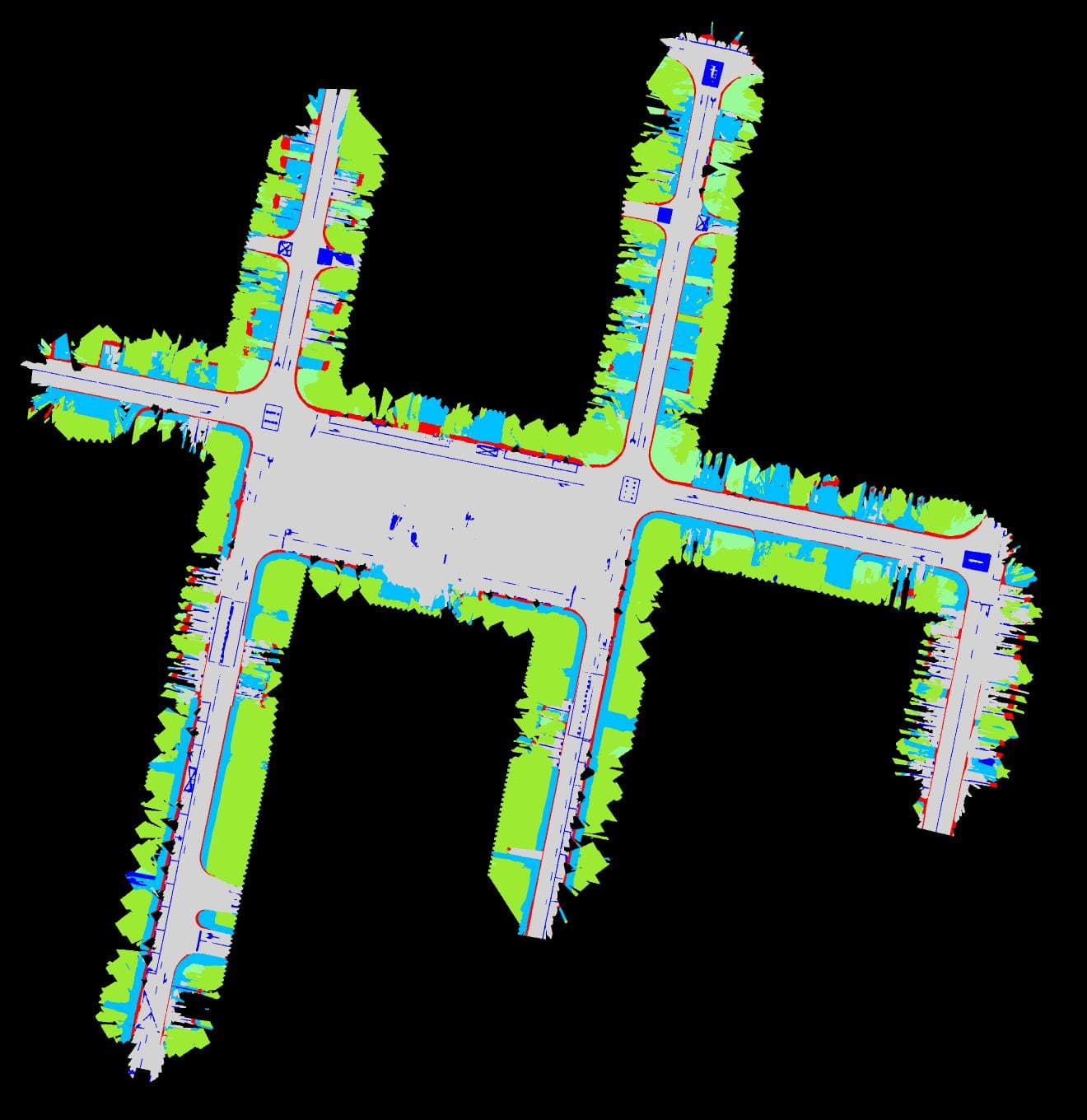}
\end{subfigure}
\\
&
\begin{subfigure}[b]{0.15\textwidth}
    \centering
    \includegraphics[width=1\linewidth]{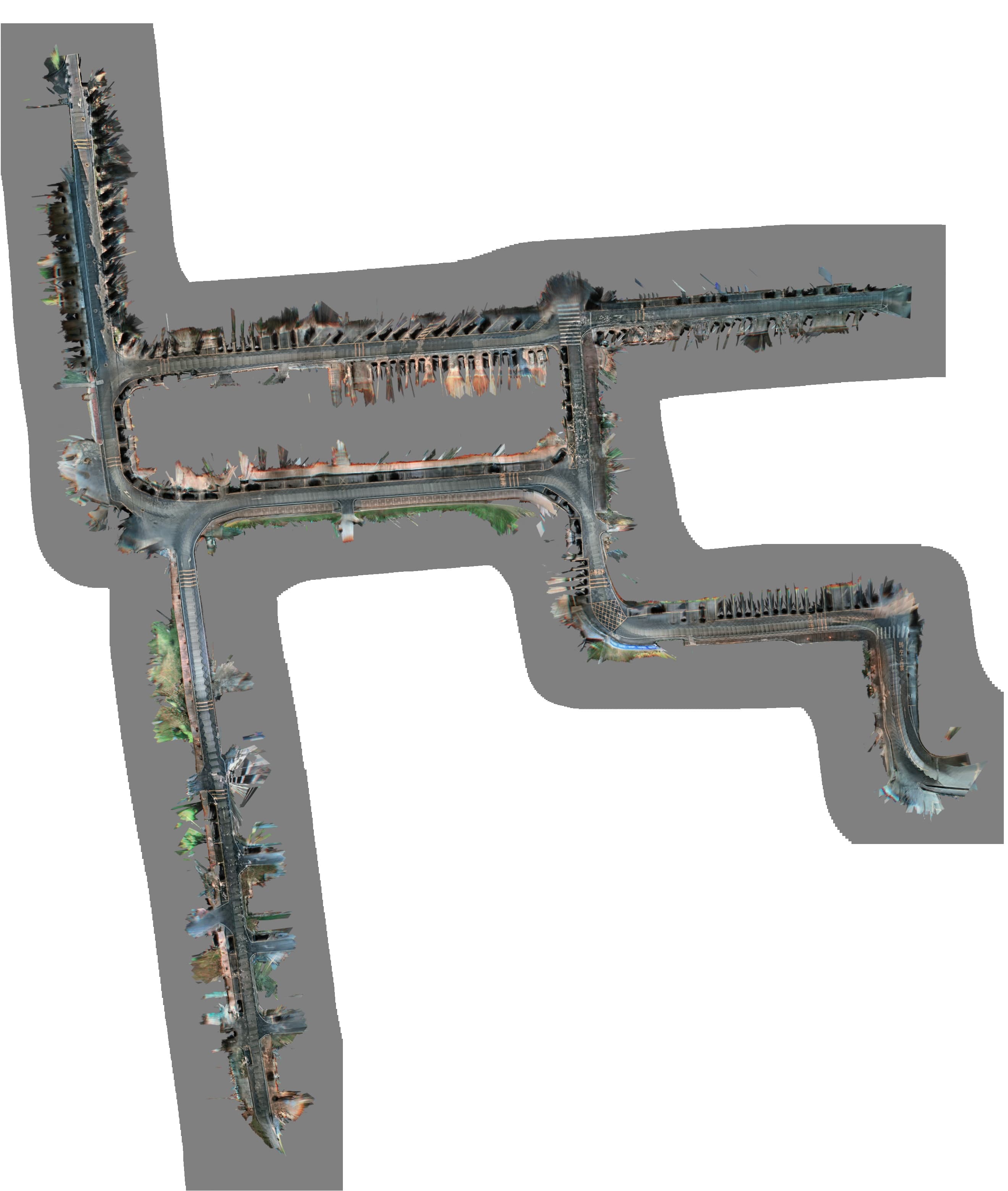}
\end{subfigure}
&
\begin{subfigure}[b]{0.15\textwidth}
    \centering
    \includegraphics[width=1\linewidth]{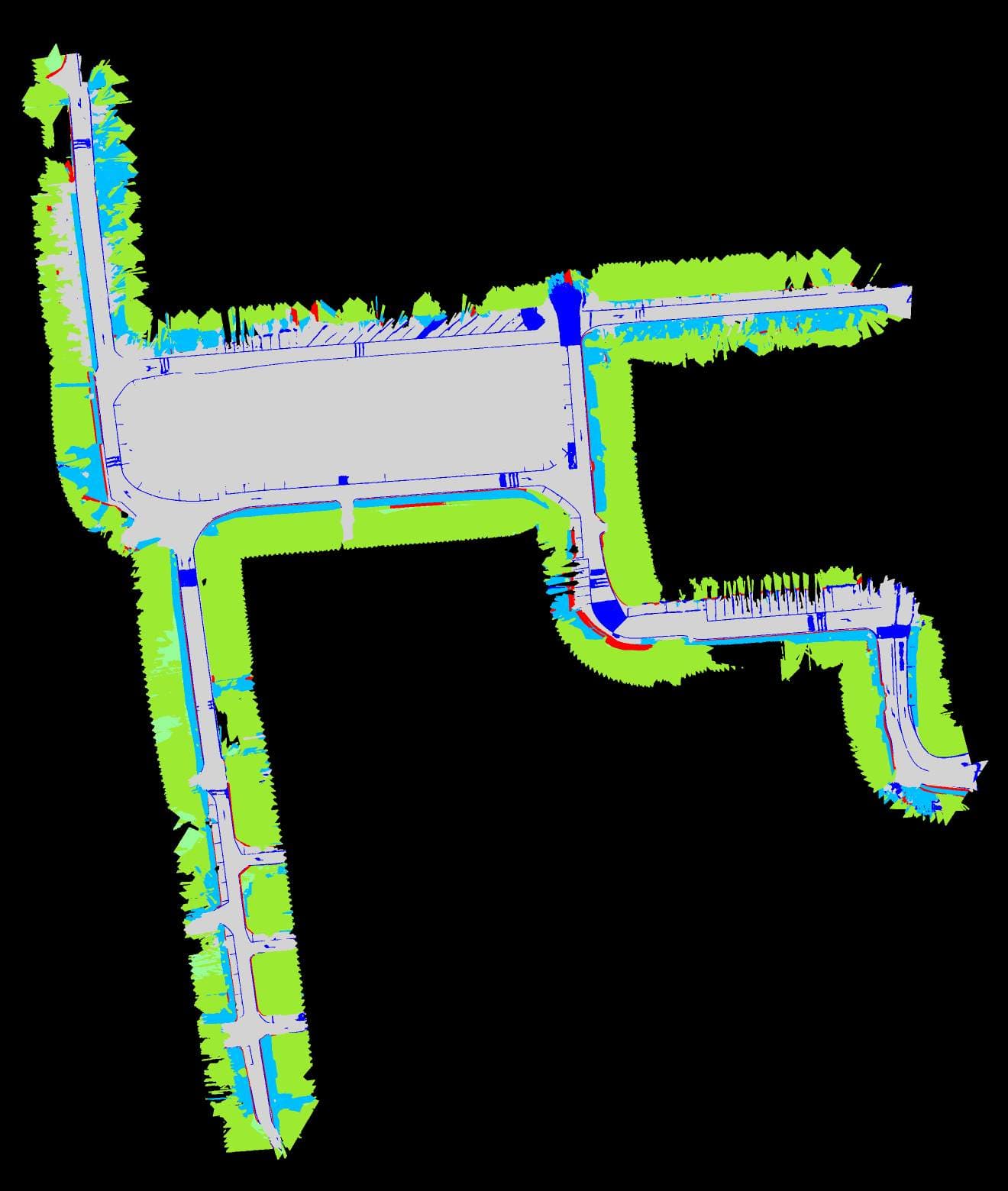}
\end{subfigure}
\\
&
\begin{subfigure}[b]{0.15\textwidth}
    \centering
    \includegraphics[width=1\linewidth]{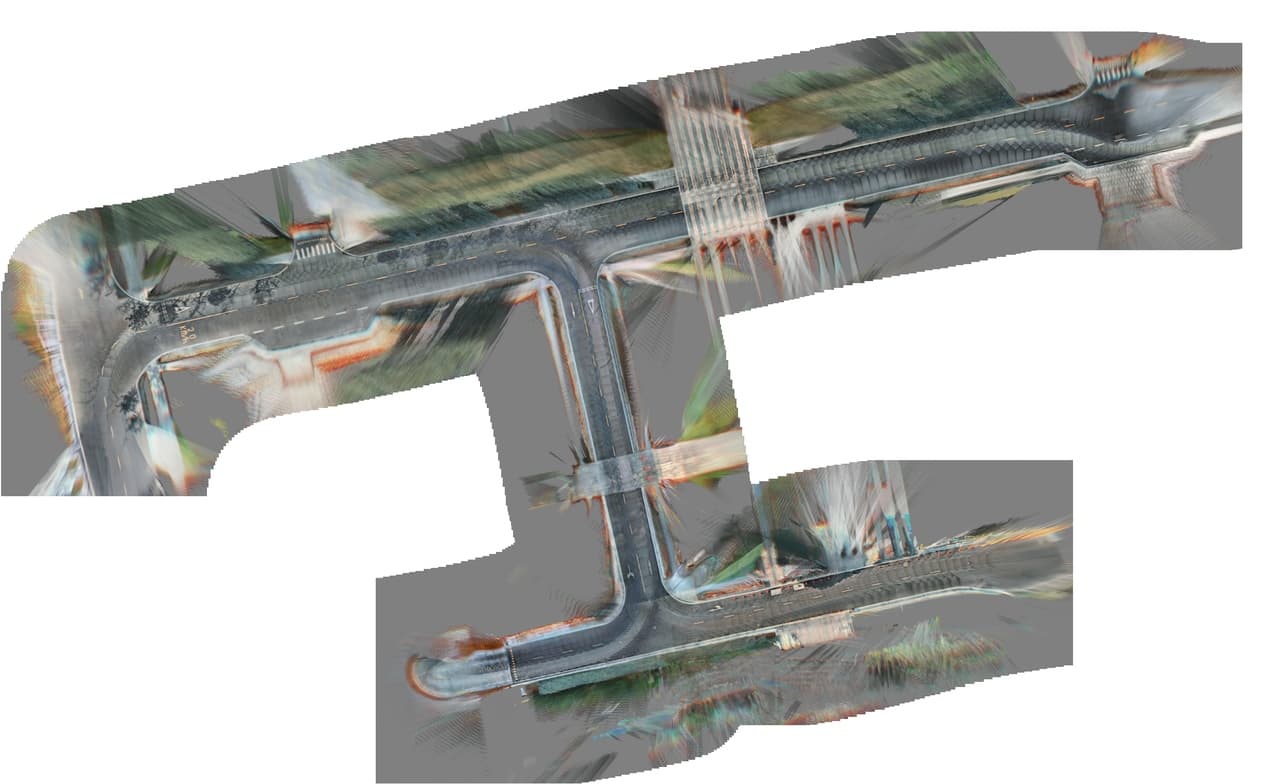}
\end{subfigure}
&
\begin{subfigure}[b]{0.15\textwidth}
    \centering
    \includegraphics[width=1\linewidth]{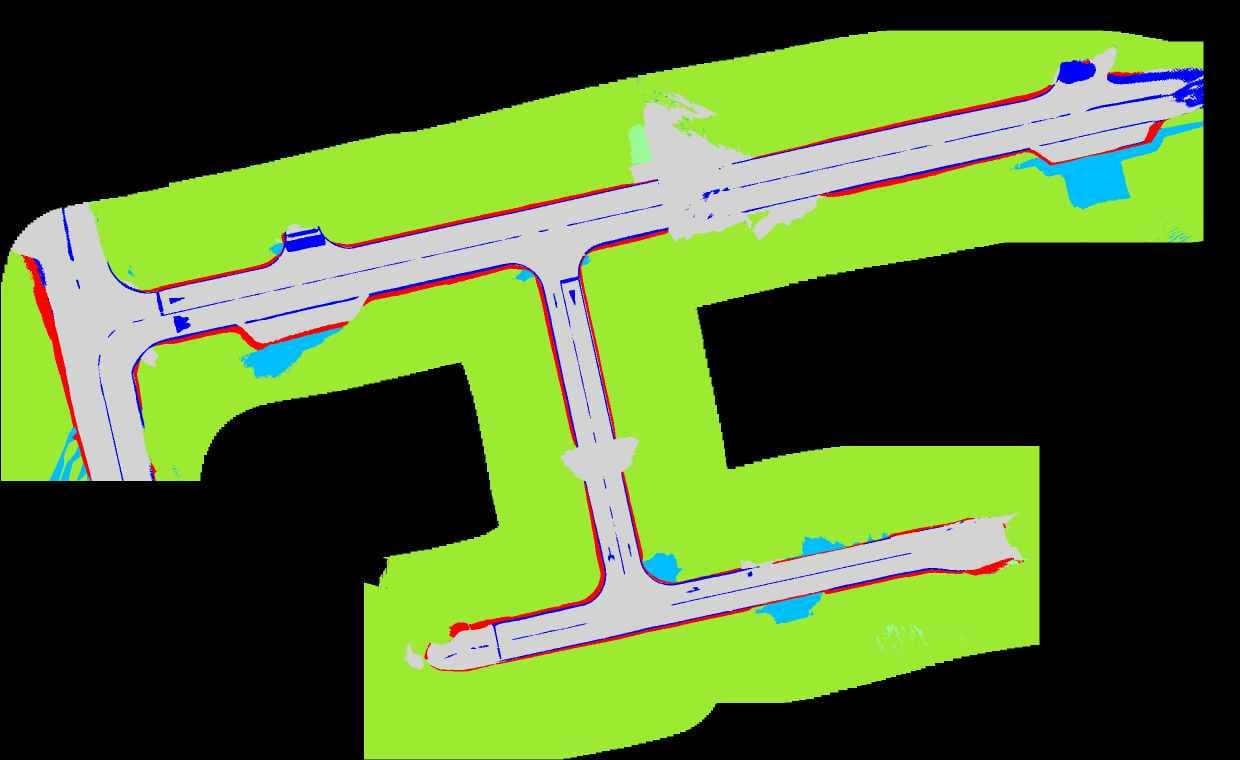}
\end{subfigure}
\\
&
\begin{subfigure}[b]{0.15\textwidth}
    \centering
    \includegraphics[width=1\linewidth]{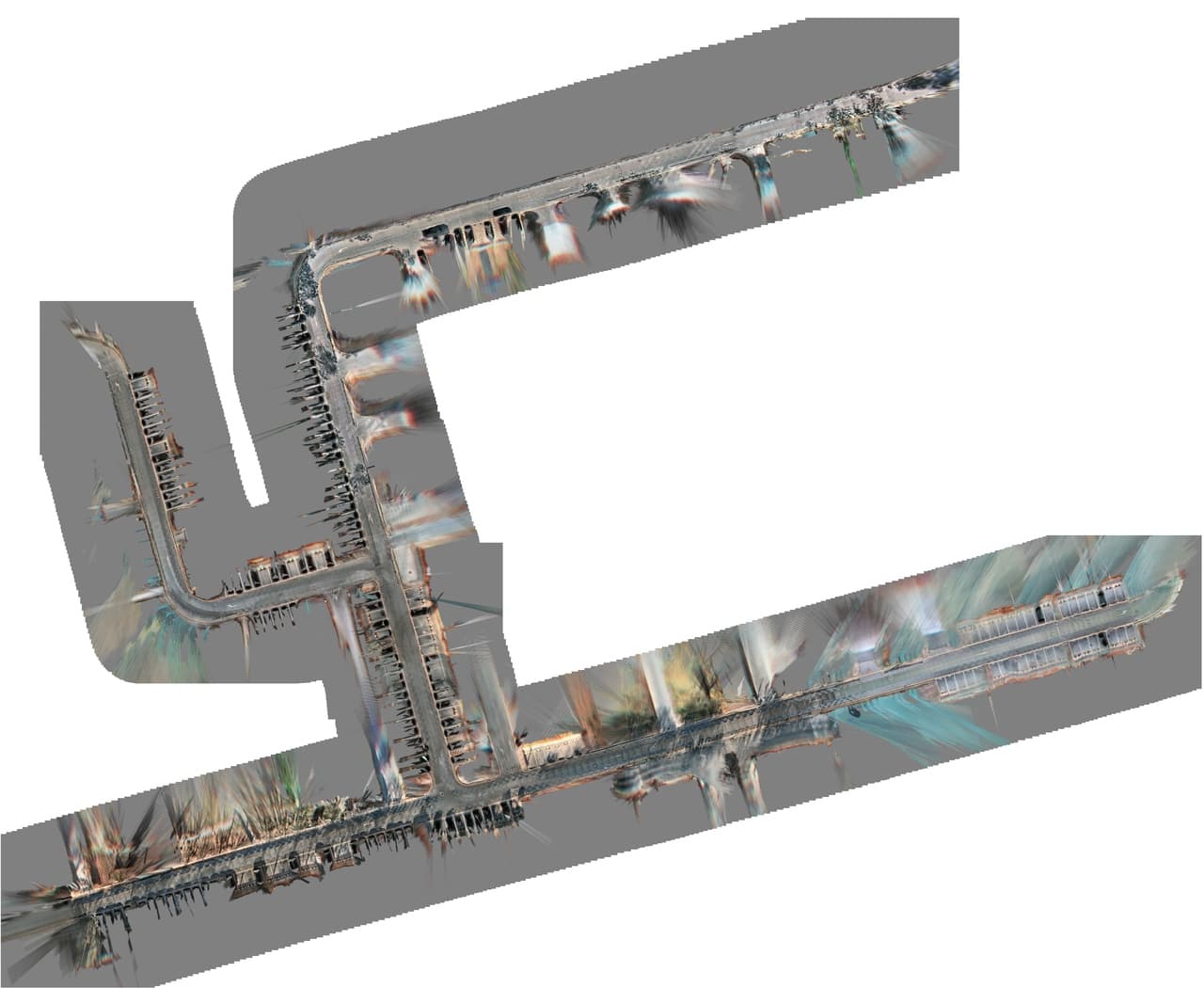}
\end{subfigure}
&
\begin{subfigure}[b]{0.15\textwidth}
    \centering
    \includegraphics[width=1\linewidth]{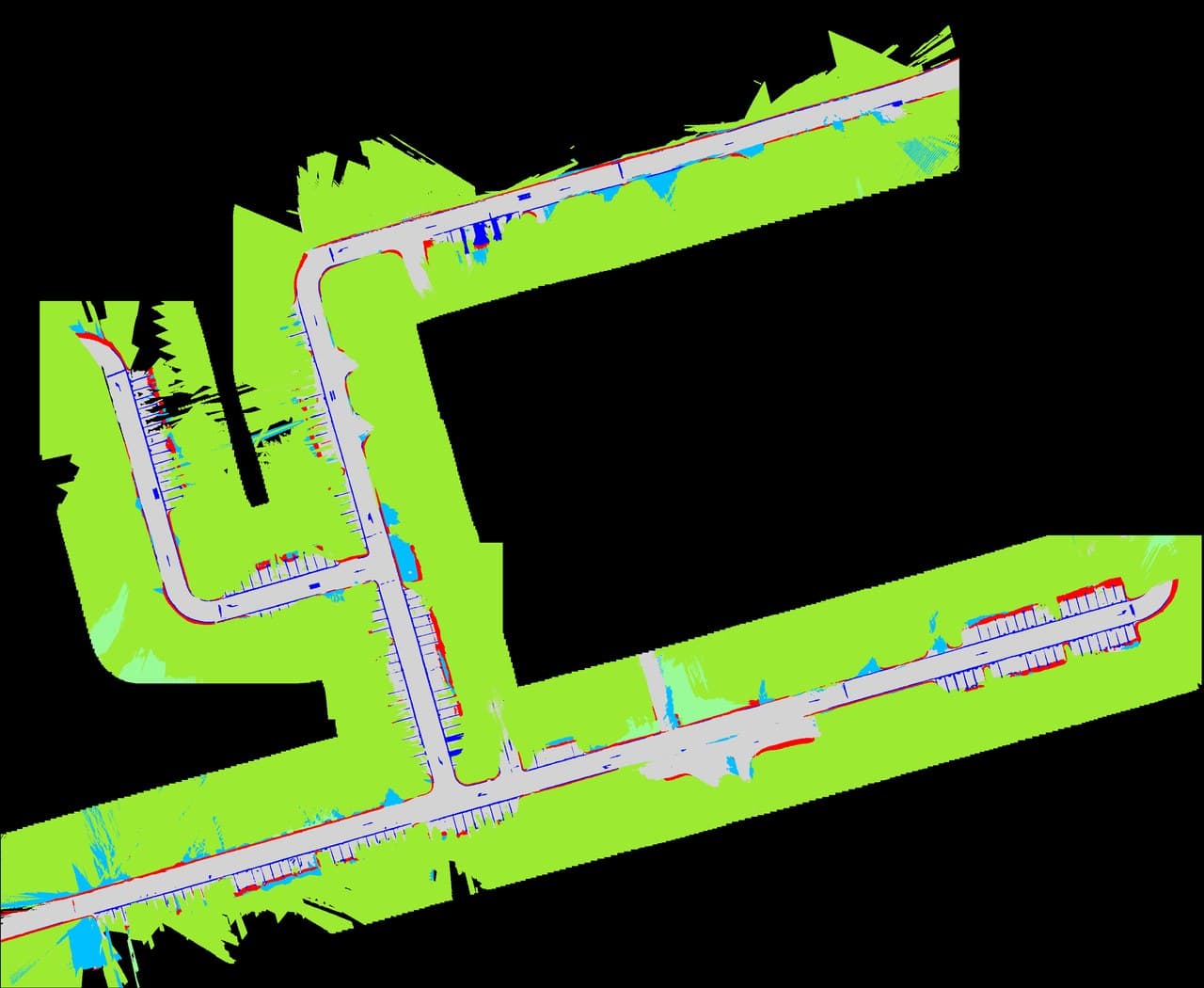}
\end{subfigure}
\\ \hline
Rural                    &
\begin{subfigure}[b]{0.15\textwidth}
    \centering
    \includegraphics[width=1\linewidth]{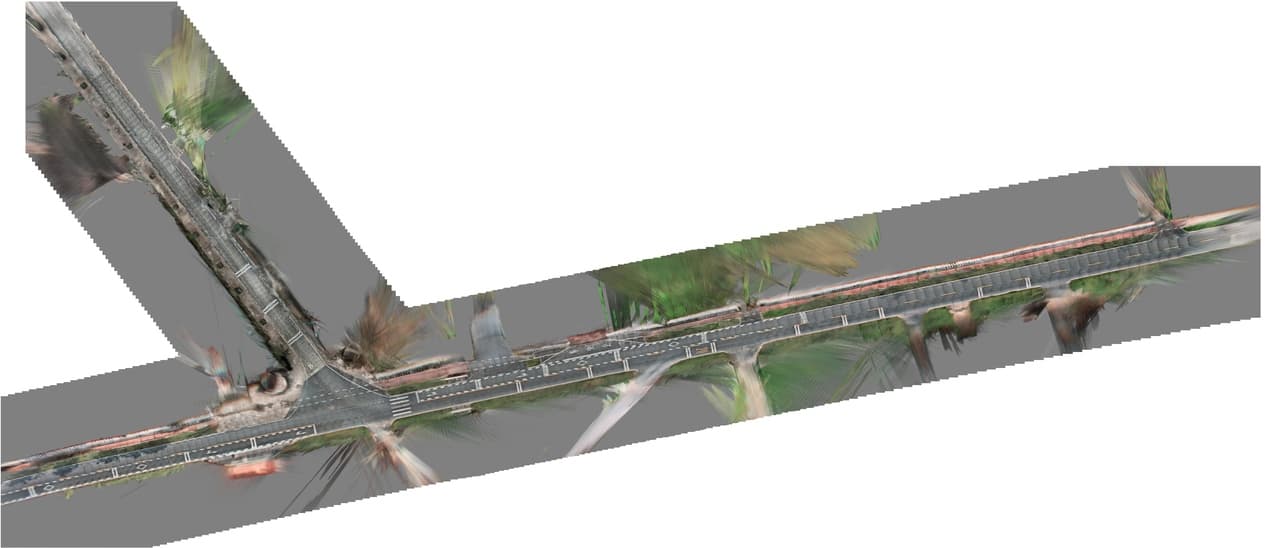}
\end{subfigure}
&
\begin{subfigure}[b]{0.15\textwidth}
    \centering
    \includegraphics[width=1\linewidth]{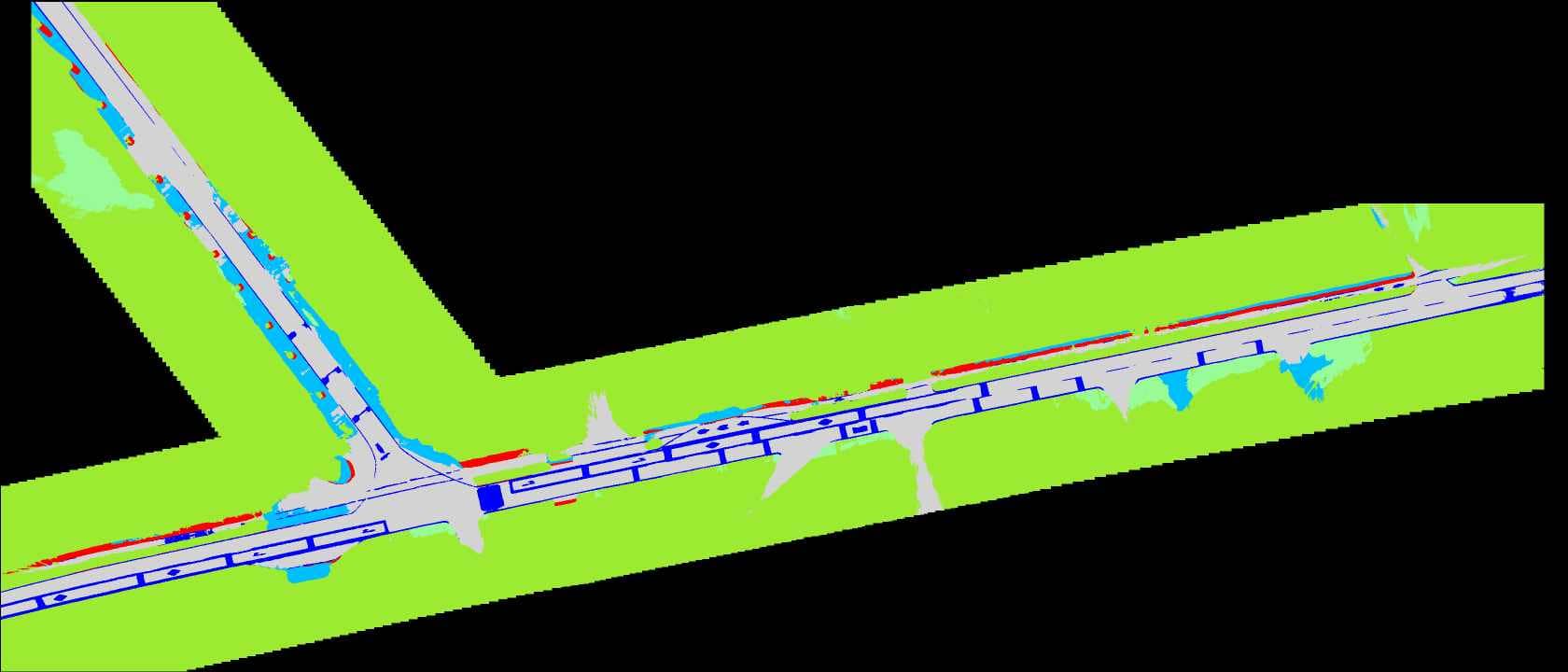}
\end{subfigure}
\\ \hline
\end{tabular}
\label{fig:bev_maps}
\end{table}

\subsection{Day/Night/Rain Distribution} 
\begin{figure}[htbp]
    \centering
    \includegraphics[width=0.45\linewidth]{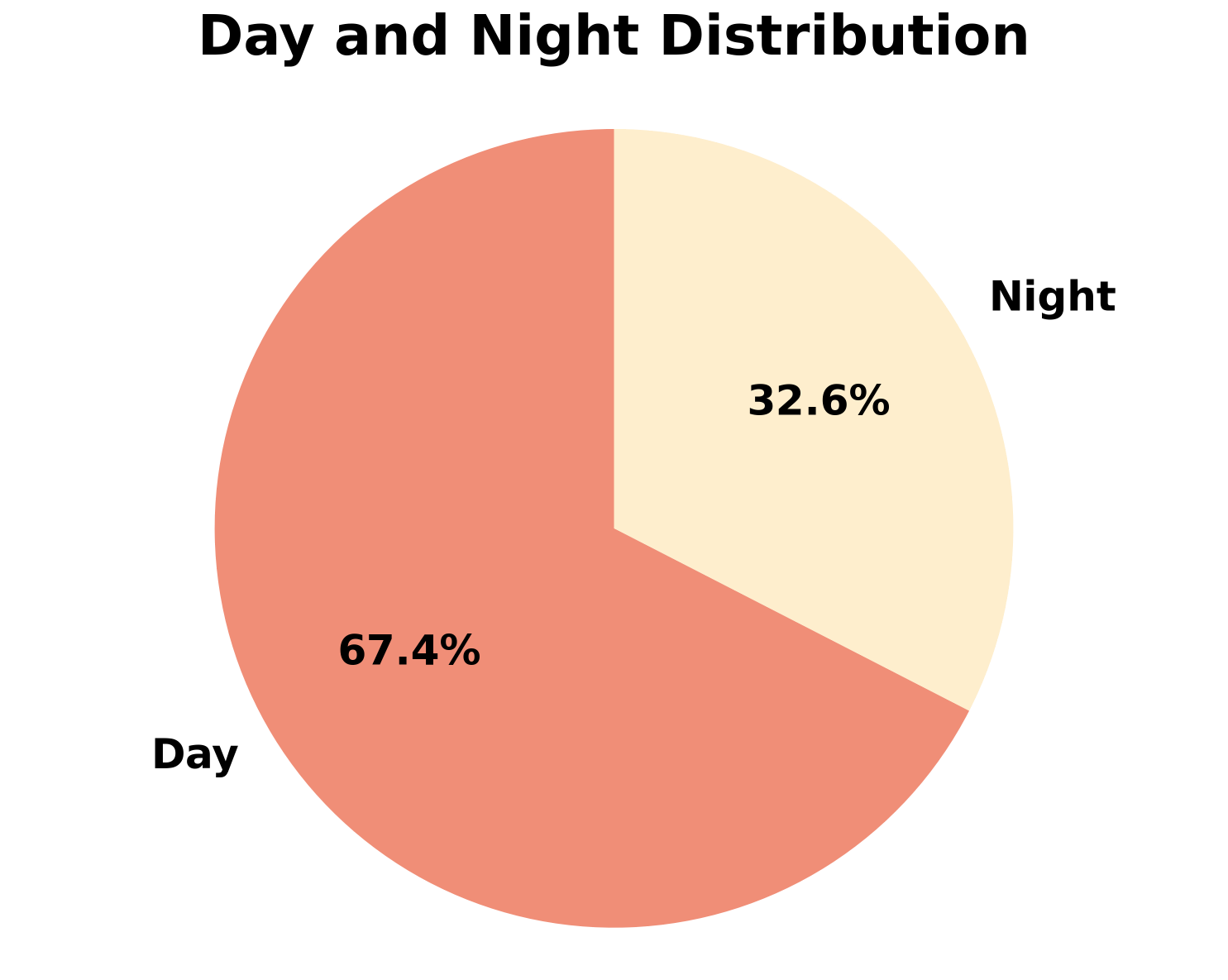}
    \includegraphics[width=0.45\linewidth]{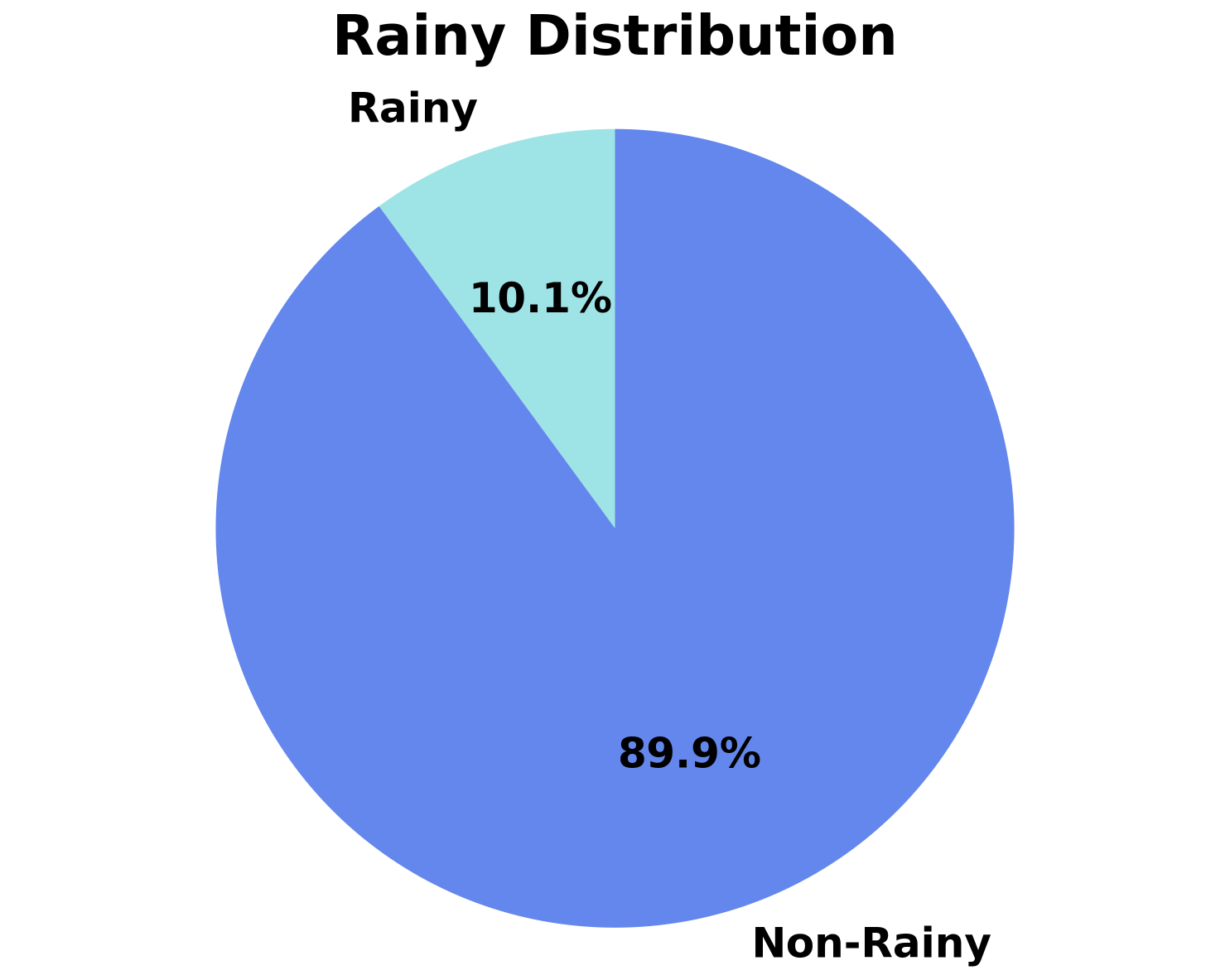}
    \caption{Distribution of day, night, and rainy conditions in our dataset. The pie chart shows the proportion of data captured under each weather and lighting condition.}
    \label{fig:day_night_rain}
\end{figure}

Our dataset is collected under a wide spectrum of environmental conditions to enhance robustness and real-world relevance. As shown in \autoref{fig:day_night_rain}, the pie chart illustrates the statistical distribution of data samples acquired during daytime, nighttime, and rainy weather. We intentionally balance the collection process to ensure substantial representation of each condition, thereby supporting the development and evaluation of algorithms that must operate reliably in varying illumination and adverse weather. This comprehensive coverage of lighting and weather scenarios is critical for building resilient perception and planning models for real-world autonomous driving.

\subsection{Dynamic Object Annotation} 
\begin{figure}[htbp]
    \centering
    \includegraphics[width=0.55\linewidth]{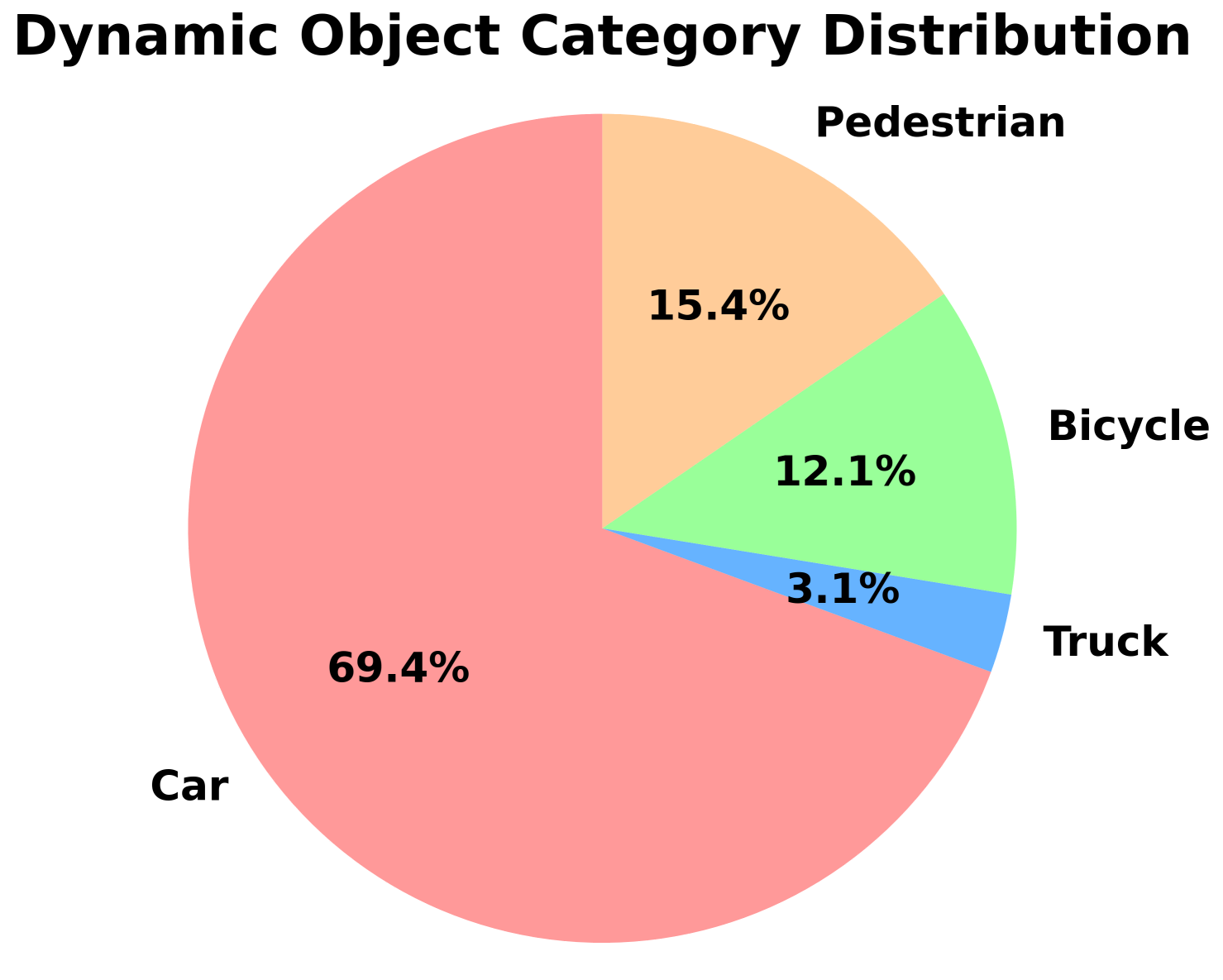}
    \caption{Distribution of dynamic object categories detected in the dataset. The pie chart presents a fine-grained breakdown of detected dynamic objects, such as cars, trucks, bicycles, and pedestrians.}
    \label{fig:dynamic_objects}
\end{figure}

To facilitate fine-grained analysis and evaluation, the dataset provides detailed 3D annotations for dynamic objects with rich category labels. \autoref{fig:dynamic_objects} presents the category-wise distribution of detected dynamic objects, including but not limited to cars, trucks, buses, motorcycles, bicycles, and pedestrians. The presence of a diverse and balanced set of object types reflects real-world traffic complexity and supports the benchmarking of detection, tracking, and scene understanding algorithms. The fine-grained categorization and abundance of dynamic object annotations further enable research on rare object detection, long-tail distribution modeling, and multi-class interaction analysis. 
The 3D bounding boxes are produced through an automated annotation pipeline based on an off-the-shelf model~\cite{liu2022petr}, making it practical to scale the dataset to diverse scenes and newly collected videos.

\subsection{Calibration Quality}

All recording vehicles are equipped with a standardized seven-camera rig. Camera calibration is performed through a three-stage pipeline, including factory calibration, online ego-vehicle calibration, and offline refinement. In the current dataset, 99.7\% ($3\sigma$) of calibration errors are below $0.3^\circ$, while the localization error remains below $0.3\%$, indicating high geometric consistency across the captured data.

\subsection{Privacy Protection}

To support responsible data release, all data will be de-identified prior to publication. In particular, personally identifiable visual content, including pedestrian faces and vehicle license plates, will be blurred before release. These measures are applied as part of the standard data processing pipeline to reduce privacy risks while preserving the utility of the dataset for research purposes.

\section{Details on Static Scene Reconstruction}
\label{supp:static_scene}
\subsection{Initialization and Block-wise Processing}
For multi-traversal image inputs, we employ MVSNet~\cite{yao2018mvsnet} to reconstruct point clouds, which are used to initialize the positions of Gaussians for the background regions. For the sky component, which is assumed to be at a considerable distance from the scene, sky Gaussians are initialized by uniformly sampling points on a hemisphere that encompasses the scene. For the ground, we initialize points on a surface fitted to the estimated ground geometry, ensuring that the Gaussians accurately capture the underlying road layout.
To model large scenes efficiently, we partition each scene into 100-meter blocks with 25\% overlap. The overlap improves continuity near block boundaries, while the block-wise strategy keeps the anchor-based background representation efficient. Buildings and other far-away structures outside the active block are absorbed into the sky node instead of being represented by a dense set of background anchors, which avoids unbounded memory growth and overly coarse distant geometry.



\subsection{Controllable Lighting Change}
\begin{figure}[htp]
    \centering
    \includegraphics[width=\linewidth]{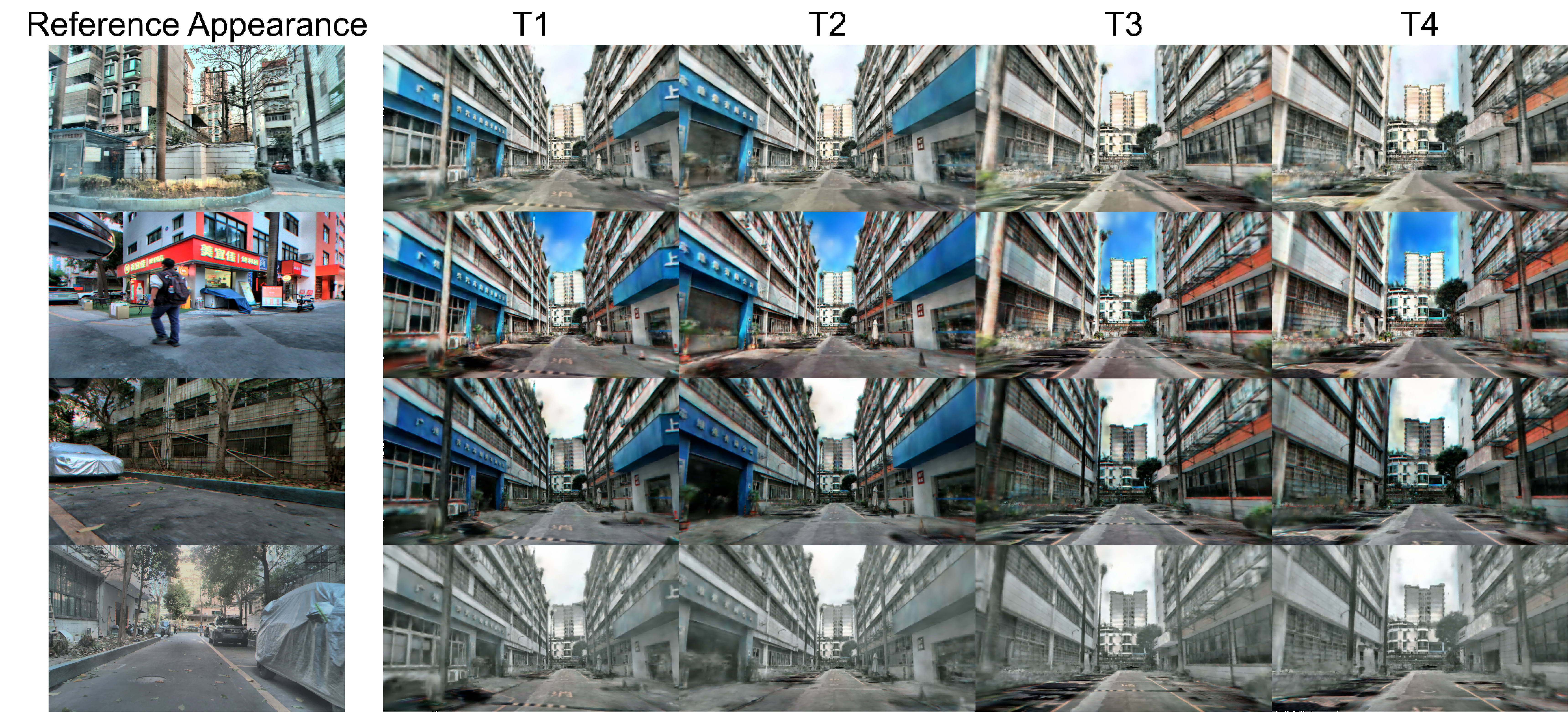}
    \caption{Temporal lighting editing.}
    \label{fig:temporal_lighting}
\end{figure}

As shown in \autoref{fig:temporal_lighting}, varying the appearance latent code produces natural changes in brightness, sky color, and shading, while keeping scene geometry unchanged across time and space. Structural elements such as roads and buildings remain stable. This demonstrates effective decoupling of scene structure and appearance, supporting realistic lighting editing for simulation.

In practice, the appearance latent code is sensor-specific and can be written as $z(j_{\text{drive}}, k_{\text{sensor}})$, where $j_{\text{drive}}$ indexes the traversal and $k_{\text{sensor}}$ indexes the camera. This design absorbs camera-dependent ISP differences while preserving traversal-level appearance control.

\subsection{Loss Function}
\label{app:loss}

During training, our objective is to ensure accurate reconstruction of both the scene appearance and geometry. To this end, we employ a combination of photometric, geometric, and normal-based constraints.

First, we supervise the rendered RGB images $I_{\text{render}}$ using the ground-truth images $I_{\text{gt}}$ with a standard photometric loss:
\begin{equation}
\mathcal{L}_{\text{rgb}} = \left\| I_{\text{render}} - I_{\text{gt}} \right\|_2^2.
\label{eq:loss_rgb}
\end{equation}
This term guides the model to faithfully reproduce visual details and color consistency with respect to the real-world observations.

In addition, we incorporate depth maps generated by MVSNet~\cite{yao2018mvsnet} as pseudo ground-truth $\hat{D}_{\text{gt}}$ for geometric supervision:
\begin{equation}
\mathcal{L}_{\text{depth}} = \left\| D_{\text{render}} - \hat{D}_{\text{gt}} \right\|_2^2.
\label{eq:loss_depth}
\end{equation}
The corresponding depth loss encourages the network to produce accurate 3D structure and spatial alignment between the rendered output and real scene geometry.

To further improve the quality of reconstructed surfaces, we introduce a normal consistency loss that enforces local geometric fidelity. Specifically, this loss is defined as the $L_1$ distance between the predicted and ground-truth normal curvature maps:
\begin{equation}
\mathcal{L}_{\text{normal}} = \left\| \mathcal{C}(\mathbf{N}_{\text{pred}}) - \mathcal{C}(\mathbf{N}_{\text{gt}}) \right\|_1.
\end{equation}
Here, $\mathcal{C}(\mathbf{N})$ denotes the curvature map derived from the normal map $\mathbf{N}$. To compute the curvature map, we first estimate spatial gradients of the normal map $\mathbf{N}$ using $3\times 3$ Sobel operators $K_x$ and $K_y$:
\begin{equation}
K_x =
\begin{bmatrix}
-1 & 0 & 1 \\
-2 & 0 & 2 \\
-1 & 0 & 1
\end{bmatrix},
\quad
K_y =
\begin{bmatrix}
-1 & -2 & -1 \\
0 & 0 & 0 \\
1 & 2 & 1
\end{bmatrix}.
\end{equation}
The gradients for each channel are computed as
\begin{align}
\nabla_x \mathbf{N} &= \mathrm{conv2d}(\mathbf{N}, K_x), \\
\nabla_y \mathbf{N} &= \mathrm{conv2d}(\mathbf{N}, K_y).
\end{align}
The curvature map is then given by
\begin{equation}
\mathcal{C}(\mathbf{N}) = \sum_{c=1}^{C} \left[ (\nabla_x \mathbf{N}_c)^2 + (\nabla_y \mathbf{N}_c)^2 \right].
\end{equation}
This process captures fine-scale surface variations and sharp features in the geometry, effectively promoting the preservation of intricate geometric details and discouraging over-smoothing artifacts.

Finally, the overall reconstruction loss combines these three components with balancing weights $\lambda_{\text{depth}}$ and $\lambda_{\text{normal}}$:
\begin{equation}
\mathcal{L}_{\text{recon}} =
\mathcal{L}_{\text{rgb}} +
\lambda_{\text{depth}} \mathcal{L}_{\text{depth}} +
\lambda_{\text{normal}} \mathcal{L}_{\text{normal}}.
\end{equation}
Here, $\mathcal{L}_{\text{rgb}}$ encourages photometric fidelity, $\mathcal{L}_{\text{depth}}$ promotes accurate geometric structure, and $\mathcal{L}_{\text{normal}}$ regularizes local surface properties, ensuring sharpness and realism in the reconstructed scenes.

\section{Details on Dynamic Scene Generation}
\label{supp:dynamic_scene}
Our dynamic generation module is built upon the layout-conditioned text-to-video (T2V) CTSD model released in OpenDWM~\cite{opendwm,chen2024unimlvg,ni2025maskgwm}, which serves as the base model in our implementation.

\subsection{Autoregressive Diffusion Objective}
To synthesize realistic and temporally coherent dynamic scenes along novel trajectories, we formulate dynamic scene generation as an autoregressive denoising process. At time step $t$, the model predicts the current target multi-view frame $I_t^{\text{out}}$ conditioned on a short history of reference frames
\begin{equation}
V_{t-T:t-1}^{\text{ref}} = \left\{ I_{t-T}^{\text{ref}}, \ldots, I_{t-1}^{\text{ref}} \right\},
\end{equation}
where $T=5$ in our implementation. During training, these reference frames are sampled from the ground-truth target sequence using teacher forcing; during inference, they are replaced by the model's previously generated outputs.

For each target frame, the static renderer first produces a target-view background image
\begin{equation}
I_t^{\text{bg}} = F_{\text{BG}}(d_t, L),
\end{equation}
where $d_t$ is the target view direction and $L$ is the appearance latent. We then encode the ground-truth target frame into a clean latent $x_0$ using a VAE, sample a diffusion timestep $s$, and add Gaussian noise $\epsilon \sim \mathcal{N}(0, I)$ to obtain the noisy latent $x_s$. The reference frames are encoded into latent features $z^{\text{ref}}$, and the rendered background image is processed by an image adapter to produce background features $f_t^{\text{bg}}$.

In addition to the latent-space inputs, the model uses auxiliary conditions
\begin{equation}
C_t = \left( B_t^{\text{ref}}, B_t^{\text{tgt}}, R_t, P_t \right),
\end{equation}
where $B_t^{\text{ref}}$ and $B_t^{\text{tgt}}$ denote the reference and target 3D layout cues, $R_t$ is the camera raymap, and $P_t$ is the text condition. The denoising network $\Phi$ is optimized with the standard noise-prediction objective
\begin{equation}
\mathcal{L}_{\text{diff}} =
\mathbb{E}_{x_0,\epsilon,s}
\left[
\left\|
\epsilon - \Phi(x_s, s, z^{\text{ref}}, f_t^{\text{bg}}, C_t)
\right\|_2^2
\right].
\label{eq:diff_loss}
\end{equation}

\subsection{Autoregressive Inference}
At inference time, we first render the static background from $F_{\text{BG}}$ using the target view and the specified appearance latent. We then form the reference window from previously generated frames, encode the window into latent features, and denoise the current latent conditioned on the rendered background and auxiliary controls. For the first $T$ time steps, the autoregressive process is initialized by seed frames; afterwards, the reference window is updated using the model's own predictions. This design decouples static geometry from dynamic content synthesis: the static model provides cross-view consistency, while the diffusion model supplies flexible dynamic generation.



\subsection{Scalability and Generalization}
Our dynamic-scene pipeline is designed to support simple and scalable integration of dynamic-object data into the simulator. In MIRROR, 3D bounding boxes of vehicles are detected automatically using PETR~\cite{liu2022petr}, and the same preprocessing pipeline can be directly applied to newly collected videos. This makes it straightforward to incorporate  unlabeled driving data and to continuously expand the simulator with new observations.

In addition, the framework shows promising generalization ability across scenes and datasets. As reported in the main paper, the dynamic results on nuScenes are obtained in an inference-only setting, without additional finetuning, indicating that the proposed pipeline can transfer effectively to unseen data.



\section{More Experiments}
\label{supp:more_results}

\begin{figure}[htp]
    \centering
    \includegraphics[width=0.95\linewidth]{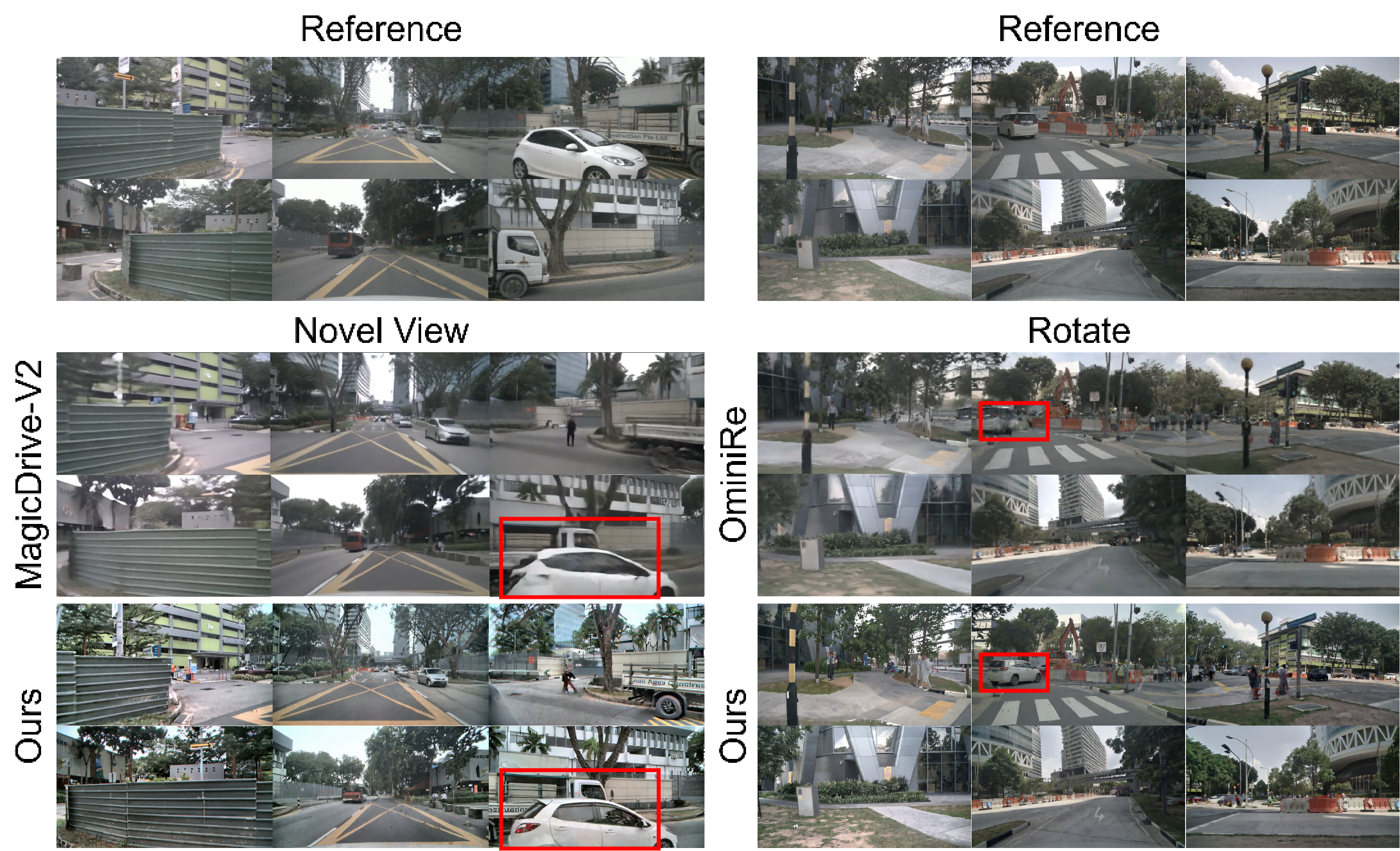}
    \caption{
    Closed-loop simulation comparisons. Left: comparison with MagicDrive-V2~\cite{gao2025magicdrive-v2} at both reference and novel views. Right: comparison with OmniRe~\cite{chen2025omnire} on dynamic vehicle editing. Our results on nuScenes are obtained in an inference-only manner without additional training.
    }
    \label{fig:comp_full}
\end{figure}

\subsection{Ablation Study}
\begin{figure}[htp]
    \centering
    \includegraphics[width=0.8\linewidth]{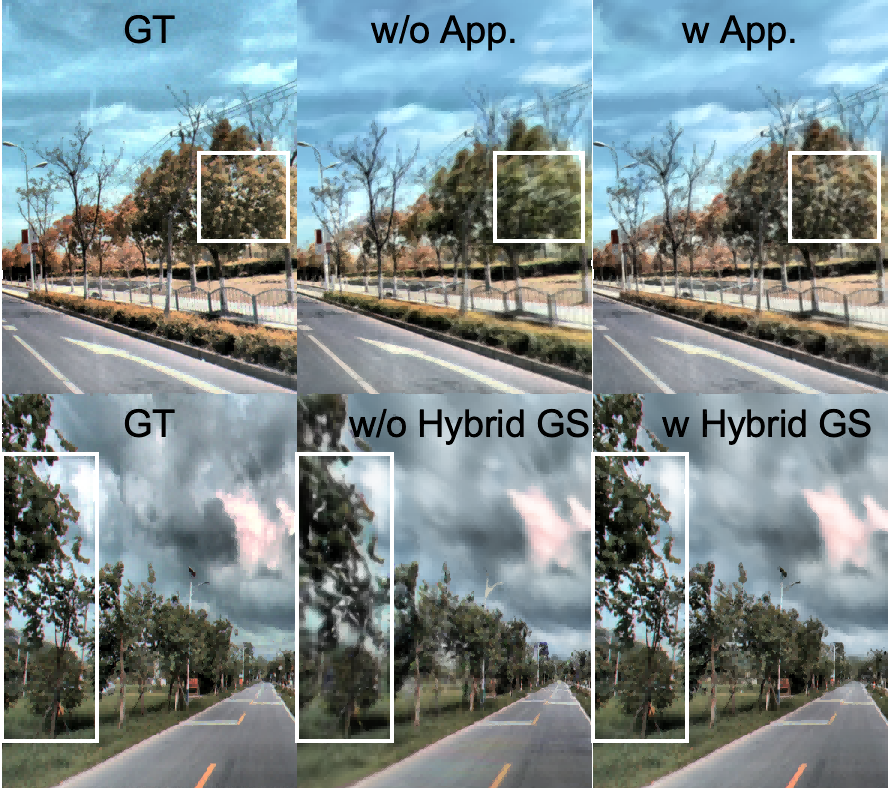}
    \caption{Ablation studies on the appearance code (App.) and hybrid Gaussian representation (hybrid GS). Here, App. denotes the appearance code, vanilla GS denotes the original 3DGS, and hybrid GS denotes our design.}
    \label{fig:ablation}
\end{figure}

We conduct ablation studies to evaluate the contributions of key components in our method. We assess the impact of the appearance latent encoding and the hybrid Gaussian representation design. \autoref{fig:ablation} shows that both the hybrid Gaussian splatting framework and the appearance latent significantly improve the modeling of transient elements, particularly in vegetation growth and seasonal color transitions. We further conduct an ablation on the dynamic generation module by comparing a baseline model without background image conditioning with our full model, which is conditioned on the target-view background image $I^{\text{bg}}$ from the static reconstruction stage. As shown in \autoref{tab:ablation}, introducing $I^{\text{bg}}$ significantly improves both FID and FVD, validating the effectiveness of target-view background conditioning for dynamic scene generation.

\begin{table}[htp]
\centering
\caption{
    Ablation on dynamic generation with background conditioning $I^{bg}$ from static reconstruction.
}
\vspace{-1em}
\resizebox{0.6\linewidth}{!}{\begin{tabular}{l|l|l}
\toprule
Setting        & FID$\downarrow$ & FVD$\downarrow$ \\
\hline
Baseline &   8.60   & 41.27
\\
+ Background Image $I^{bg}$ (Full)    &    \textbf{6.27}    & \textbf{25.51}                 \\
\bottomrule
\end{tabular}}
\label{tab:ablation}
\end{table}

\subsection{Additional Reconstruction Comparisons}
\begin{figure}[H]
    \centering
    \includegraphics[width=\linewidth]{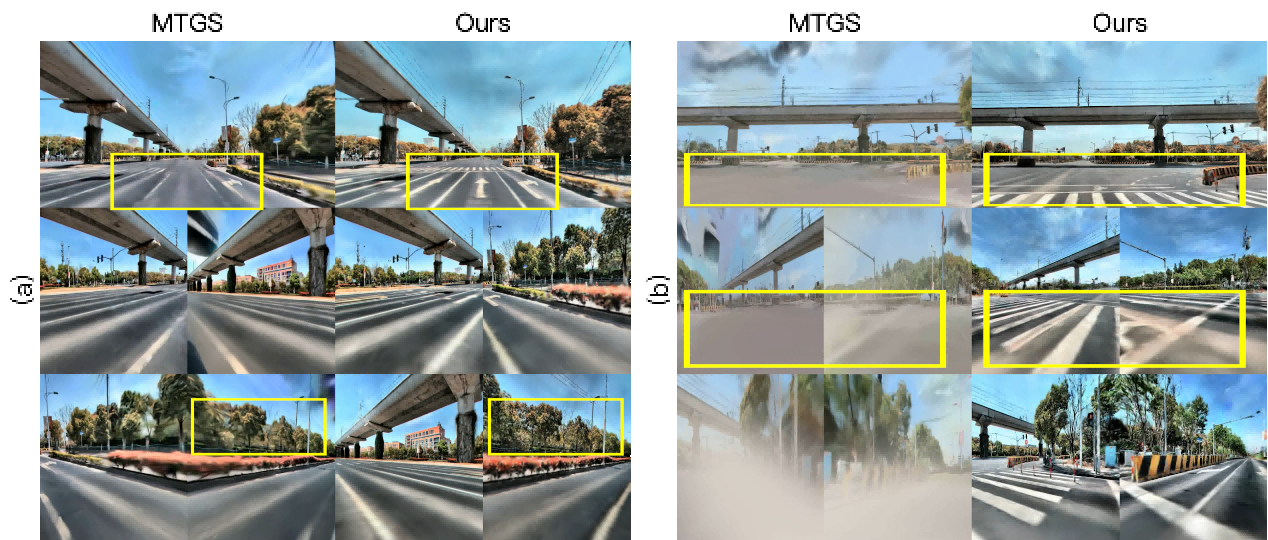}
    \caption{Comparison of our static scene reconstruction module with MTGS~\cite{li2025mtgs} on the MIRROR dataset.}
    \label{fig:supp_recon}
\end{figure}

The nuPlan results reported in the main paper are averaged over all six road blocks rather than a selected subset. We further compare our approach with existing baselines to highlight its advantages in reconstruction quality and editing stability. As shown in \autoref{fig:supp_recon}, we provide additional evaluations between our static scene reconstruction module and MTGS~\cite{li2025mtgs} on the MIRROR dataset. In (a), our method recovers finer details in lane markings and vegetation structure compared to MTGS. In (b), under extreme viewpoint variations, MTGS fails to reconstruct plausible road geometry, while our approach maintains accurate road layout and clear traffic signs.

\begin{figure*}[htp]
    \centering
    \includegraphics[width=1.0\textwidth]{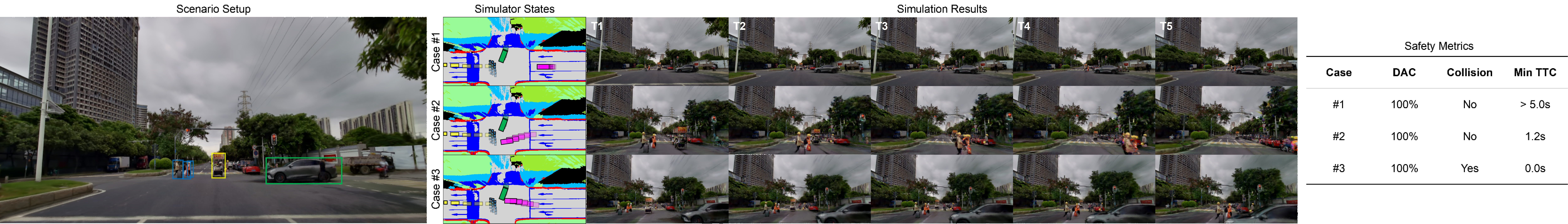}
    \caption{Safety-oriented analysis under controlled scenarios.}
    \label{fig:safety}
\end{figure*}

\subsection{Additional Scene Editing Examples}
\begin{figure}[H]
    \centering
    \includegraphics[width=0.9\linewidth]{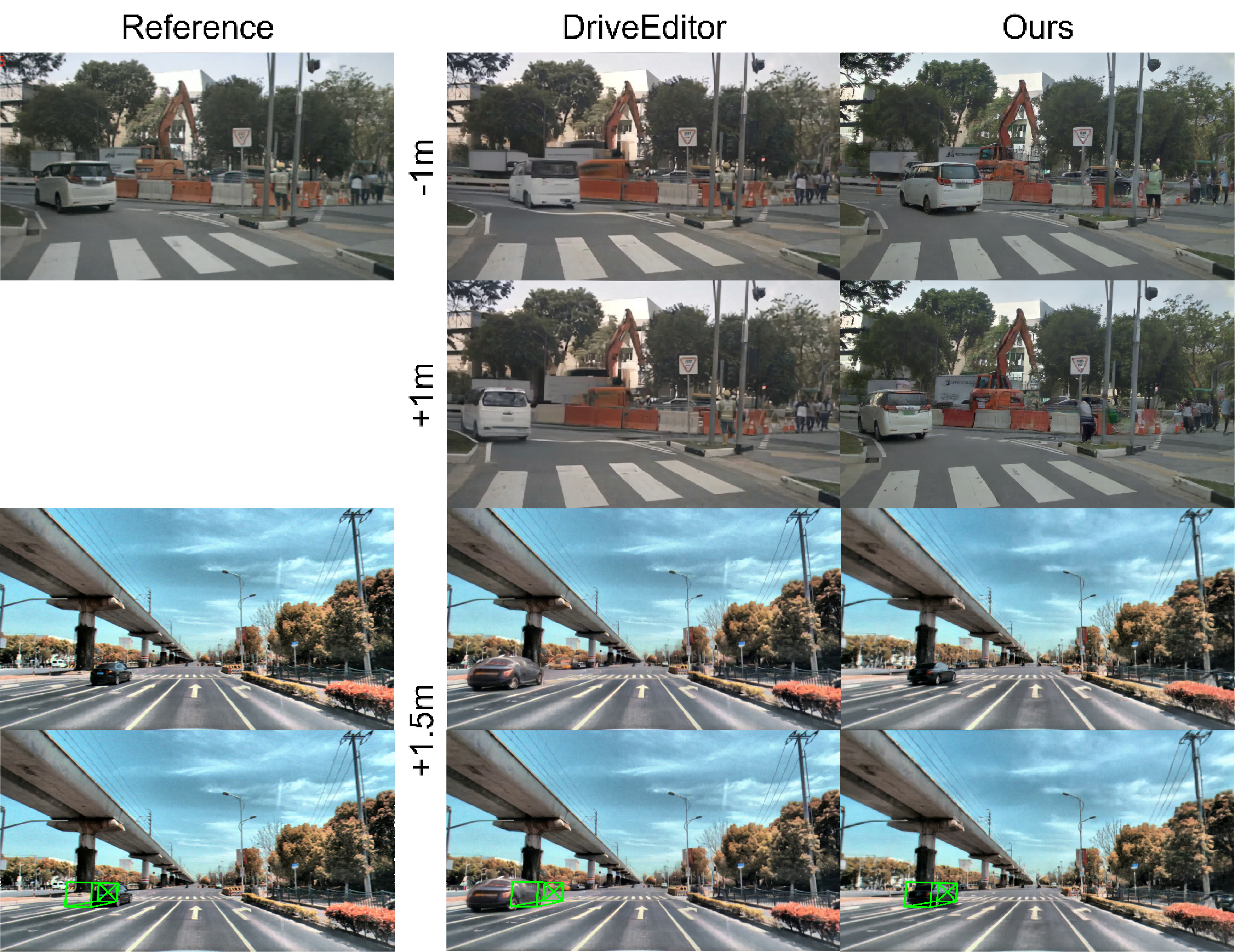}
    \caption{
    Controlled vehicle translation results.
    We compare with DriveEditor~\cite{liang2024driveeditor} under horizontal vehicle offsets and visualize the projected bounding boxes at the target positions.
    }
    \label{fig:edit}
\end{figure}

To complement the quantitative vehicle translation benchmark reported in the main paper, \autoref{fig:edit} provides qualitative comparisons with DriveEditor under different horizontal offsets. Our framework produces vehicles with more reasonable scale and more seamless background integration while maintaining accurate placement at the target positions.

\begin{figure}[H]
    \centering
    \includegraphics[width=\linewidth]{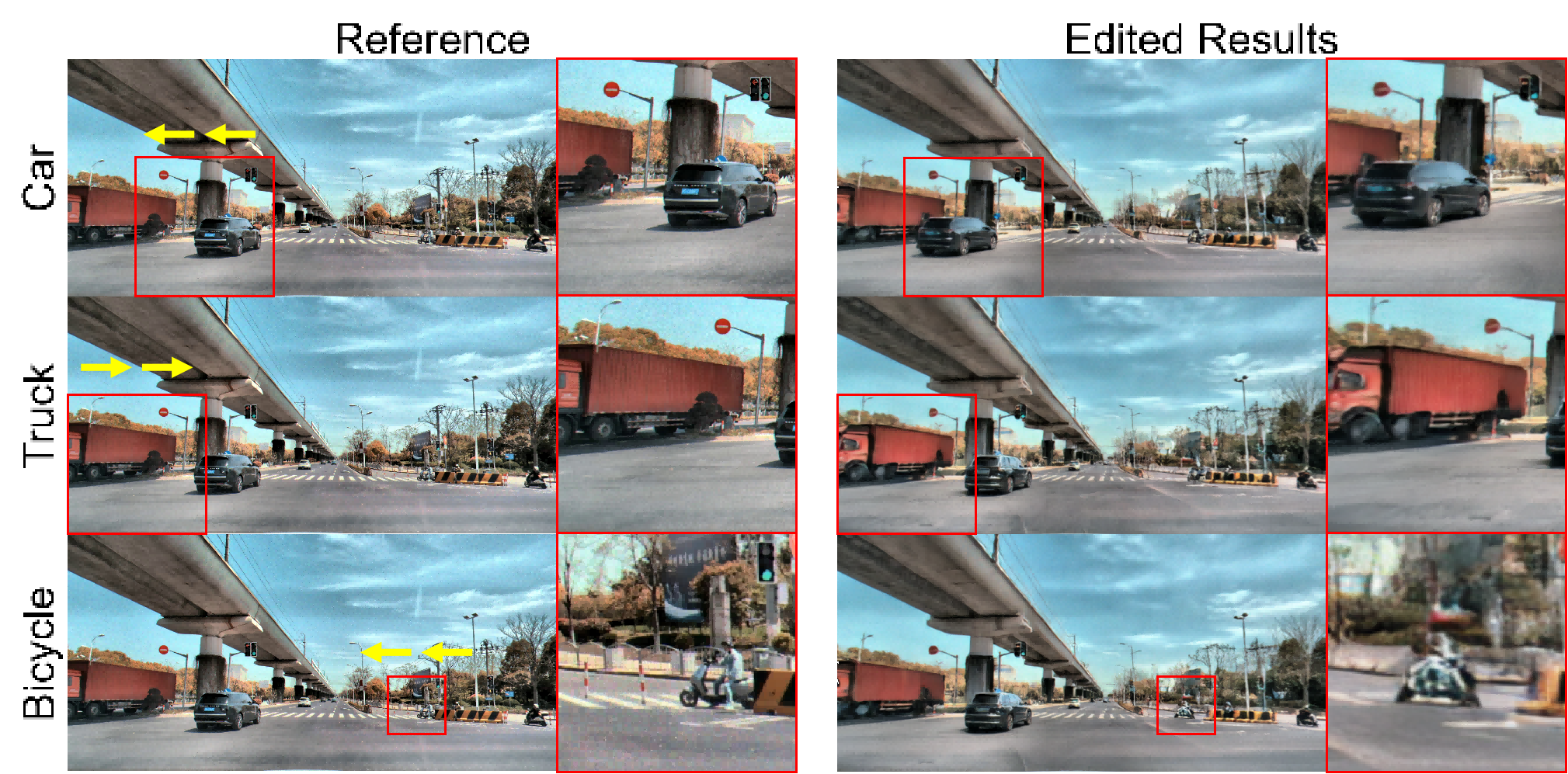}
    \caption{Additional scene-editing results. The box-based control interface supports diverse agents edits, including car, truck, and bicycle editing.}
    \label{fig:supp_edit_agents}
\end{figure}

Our framework supports scene extension without retraining: newly captured route data can be used as reference videos, modified trajectories can be specified directly, and edited videos can be synthesized immediately. Beyond controlled vehicle translation, \autoref{fig:supp_edit_agents} shows representative examples of vehicle and pedestrian editing, demonstrating that the proposed box-based control generalizes to a broader set of scene-editing operations.

We also provide additional comparisons with DriveEditor~\cite{liang2024driveeditor} in the supplementary video. The MIRROR dataset poses challenges such as sparse temporal sampling and large inter-frame displacements due to realistic driving speeds. In these scenarios, DriveEditor often produces unstable results and visible artifacts, while our method maintains stronger spatiotemporal coherence and more accurate scaling of dynamic objects.

\subsection{Editing Accuracy}
To evaluate instruction following more directly, we additionally use a SAM-based 2D evaluation protocol. We first extract the edited object region from the generated result and then compute the IoU between this region and the projected target 3D box in the image plane. As shown in \autoref{tab:edit_iou}, our method achieves substantially higher editing accuracy than DriveEditor.

\begin{table}[htbp]
    \centering
    \caption{Editing accuracy measured by IoU between the SAM-extracted edited object region and the projected target box. Higher is better.}
    \label{tab:edit_iou}
    \setlength{\tabcolsep}{10pt}
    \renewcommand{\arraystretch}{1.05}
    \begin{tabular}{lc}
        \toprule
        Method & IoU $\uparrow$ \\
        \midrule
        DriveEditor & 0.29 \\
        Ours & \textbf{0.71} \\
        \bottomrule
    \end{tabular}
\end{table}

\subsection{Safety-oriented Scenario Analysis}
HybridWorldSim provides the essential simulation capabilities required for downstream autonomous driving evaluation. Specifically, our framework supports trajectory-conditioned rollout, synchronized multi-camera observations, dynamic agent editing, and maintained scene states with ego poses and actor
bounding boxes.

To demonstrate these capabilities, we perform a safety-oriented scenario analysis using controlled scene modifications. We simulate representative safety-critical events, including vehicle crossing and dynamic object interactions, and roll out the modified scenes over multiple time steps. Based on the maintained BEV states, we
compute commonly used safety indicators, including collision occurrence, minimum time-to-collision (TTC), and drivable-area compliance (DAC).

\autoref{fig:safety} summarizes the results. These analyses verify that HybridWorldSim can generate controllable scenarios and provide the necessary state information.

\subsection{Runtime and Model Size}
We report additional efficiency statistics in \autoref{tab:runtime}. The average size of the static scene model is about 400\,MB for a 100-meter region. Static-only rendering runs at 21 FPS at $960\times 540$.

\begin{table}[htbp]
    \centering
    \caption{Additional runtime and memory statistics.}
    \label{tab:runtime}
    \setlength{\tabcolsep}{6pt}
    \renewcommand{\arraystretch}{1.05}
    \begin{tabular}{lc}
        \toprule
        Item & Value \\
        \midrule
        Static model size (100\,m region) & 400\,MB \\
        MVSNet initialization speed & 8 FPS @ $1280\times 960$ \\
        Static-only rendering speed & 21 FPS @ $960\times 540$ \\
        \bottomrule
    \end{tabular}
\end{table}

\end{document}